\documentclass[11pt]{article}
\usepackage[letterpaper,margin=1in,footskip=0.6in]{geometry}
\usepackage[T1]{fontenc}
\usepackage{times}
\usepackage{microtype}
\usepackage[round,authoryear]{natbib}
\setcitestyle{citesep={;},aysep={,},yysep={;}}

\usepackage{comment}
\usepackage{xcolor}
\usepackage{caption}
\usepackage{wrapfig}
\usepackage{pgfplots,wrapfig}
\pgfplotsset{compat=1.18}
\usepackage{amsmath,amssymb,amsthm,bm}
\usepackage{graphicx,booktabs,array,multirow}
\usepackage{xcolor,tikz,pgfplots}
\usetikzlibrary{arrows.meta,positioning, fit, backgrounds, calc,patterns, shapes.geometric}
\pgfplotsset{compat=1.18}
\usepackage[most]{tcolorbox}
\usepackage{hyperref,url}

\hypersetup{colorlinks=true,linkcolor=blue!45!black,citecolor=blue!45!black,urlcolor=blue!45!black}

\newcommand{\R}{\mathbb R}
\newcommand{\one}{\mathbf 1}
\newcommand{\diag}{\operatorname{diag}}

\newcommand{\loss}{\mathcal L}

\definecolor{foam}{HTML}{167D8D}
\definecolor{gs}{HTML}{B66A1E}
\definecolor{classa}{HTML}{276BB1}
\definecolor{classb}{HTML}{D58A25}
\definecolor{ink}{HTML}{25364A}
\tikzset{>=Latex,every node/.style={font=\small},
  panel/.style={draw=ink!25,rounded corners=2pt,fill=ink!2},
  flow/.style={-{Latex[length=2mm]},draw=ink,line width=.8pt}}
\usepackage{longtable}

\title{OV-Foam: How 3D Representations\\Shape Semantic Feature Lifting}
\author{Yousef Rajeh$^{1,*}$\quad Alejandra Esparza Pelcastre$^{2}$\quad Ivan Viola$^{1}$\quad Dominik Engel$^{1}$\\[4pt]
{\normalsize $^{1}$King Abdullah University of Science and Technology (KAUST)\quad $^{2}$Universidad Panamericana}\\
{\small \texttt{\{yousef.rajeh,\,ivan.viola,\,dominik.engel\}@kaust.edu.sa}\quad \texttt{0263483@up.edu.mx}}\\
{\small $^{*}$Corresponding author}}
\date{}
\begin{document}
\maketitle

\begin{abstract}
Existing feature lifting approaches for open-vocabulary segmentation rely on simple backprojection to assign features to primitives. In the case of 3D Gaussian splatting reconstruction, the primitives overlap heavily, leading to suboptimal reconstructions, governed by many primitives per ray. Backprojection, however, reduces to a least-squares solvable problem, exactly when no two primitives share a ray. Motivated by this insight, we propose OV-Foam, which uses PowerFoam as the underlying representation. This representation uses a power-diagram to partition the scene, resulting in zero overlap between primitives and a substantial reduction in primitives per ray, which ultimately simplifies the lifting problem and allows for closer-to-optimal feature assignment. We show that the space partitioning primitives of OV-Foam reach near-optimal assignment, and our experiments demonstrate improved open-vocabulary segmentation over Gaussian baselines.
\end{abstract}
\begin{figure}[!h]
\centering\resizebox{\linewidth}{!}{\ignorespaces\input{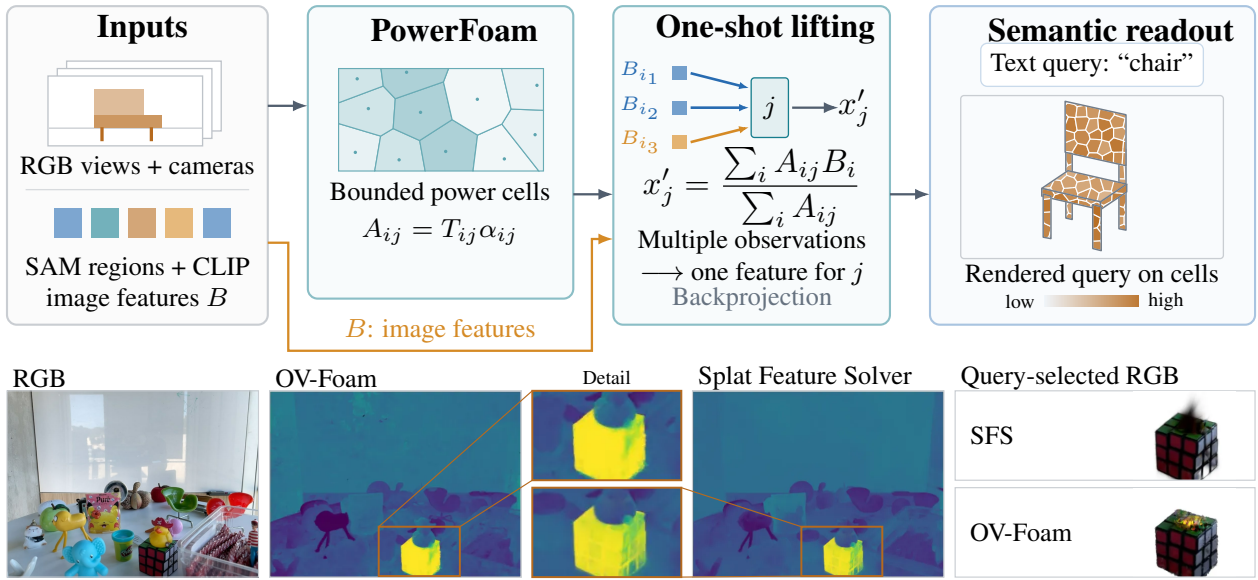}\unskip}
\caption{\textbf{Pipeline overview and qualitative example.} Starting from the input views, we create SAM masks, which are fed to CLIP to create image features \textit{$B$}. The views also reconstruct our space-partitioning PowerFoam scene, of which we then use the rendering weights \textit{$A$} for the back projection and assign each primitive a queryable feature \textit{$x'$} that is compared to the text features.}
\label{fig:teaser}
\end{figure}
\section{Introduction}
\label{sec:intro}

Open-vocabulary 3D segmentation assigns vision-language aligned semantic features, such as those produced by CLIP~\citep{radford2021learning},  to reconstructed scenes, allowing objects and spatial regions to be identified using free-form language queries. This capability supports applications such as robotic manipulation~\citep{gu2024conceptgraphs}, complex scene understanding~\citep{kerr2023lerf} and medical-imaging~\citep{wang2025sam}. Existing approaches include grouping-based methods, training-based methods, and heuristic-based methods. We focus on the latter family, which directly lifts features from 2D images onto a 3D representation.

The process of assigning these features to scene geometry is known as \emph{feature lifting}. It associates features observed in rendered images with the primitives contributing to each pixel, using their rendering weights to aggregate information across views. This provides an efficient way to obtain semantic representations without training a 3D foundation model from scratch, but introduces a fundamental ambiguity: a 2D observation may be distributed across multiple primitives, while a primitive may collect observations from different objects. As a result, feature lifting can entangle semantic evidence before a classifier makes a prediction.

This ambiguity is particularly important near object boundaries. Repeating the lifting process across views can produce feature entanglement and degraded semantic boundaries, even when the underlying image features are accurate. Recent work has begun to formalize this problem by treating feature lifting as a linear inverse problem and quantifying the ambiguity introduced by multiple primitive contributions~\citep{xiong2026sfs}.

Existing feature-lifting methods largely operate on Gaussian-based reconstructions, where overlap leads to multiple primitives contributing to the same ray. This raises a more fundamental question: to what extent is feature-lifting ambiguity a property of the lifting procedure itself, and to what extent is it induced by the underlying 3D representation?

In this work, we investigate feature lifting as a \emph{representation-dependent} problem. We ask whether changing the geometric primitives used to represent the scene can reduce the mixing of semantic evidence, while keeping the image features and lifting procedure fixed. We consider PowerFoam~\citep{govindarajan2026power}, which represents a scene using bounded power diagrams whose cells partition 3D space. This non-overlapping spatial support reduces the number of primitives that can contribute simultaneously to a ray, reducing semantic feature mixing.

We isolate this effect, by evaluating feature lifting independently of errors introduced by the upstream 2D semantic model. We replace predicted image semantics with ground-truth segmentation and text features, allowing us to measure the extent of semantic feature mixing caused by a feature lifting onto a chosen underlying 3D representation. Under this controlled setting, we compare Gaussian-based and PowerFoam reconstructions using the same lifting procedure and find that the reduced dispersion in PowerFoam leads to substantially more class-consistency in lifted features. This demonstrates that the choice of 3D primitive can itself impose a significant limitation on semantic feature lifting. Applied to real datasets, we observe that all our tested lifting methods achieve improved 3D open-vocabulary segmentation results using PowerFoam instead of 3DGS as underlying representation.

\enlargethispage{2\baselineskip}
Our contributions are:
\begin{itemize}
\item \textbf{A controlled study of representation-dependent feature lifting.} We isolate the feature-lifting stage from upstream semantic errors and systematically compare 3D primitives under the same image features and lifting procedure.
\item \textbf{An analysis of representation-induced feature dispersion.} We show that space-partitioning primitives can substantially reduce semantic feature mixing compared with overlapping Gaussian primitives.%
\item \textbf{An open-vocabulary 3D segmentation system based on PowerFoam.} We incorporate these insights into a PowerFoam-based semantic representation and demonstrate improved performance on open-vocabulary 3D segmentation benchmarks.
\end{itemize}

\section{Related work}
\label{sec:related}

Existing approaches for lifting 2D features to 3D primitives differ primarily in how the correspondence between image observations and 3D primitives is established: through training-based, grouping-based, and heuristic-forward methods.

Early approaches fit continuous 3D semantic fields via scene-specific optimization. LERF, for example, distills CLIP features into a radiance field~\citep{kerr2023lerf, radford2021learning}, while LangSplat trains a scene-specific autoencoder to represent lifted features in a latent space~\citep{qin2024langsplat}. OpenGaussian similarly learns instance-level features together with a codebook for point-level understanding~\citep{wu2024opengaussian}. These methods can produce expressive semantic representations, but  require computationally expensive per-scene optimization and may inherit inconsistencies from the underlying multi-view 2D predictions.

An alternative is replacing continuous optimization with discrete associations derived from 2D masks (e.g., SAM~\citep{kirillov2023sam}). %
LAGA decomposes scenes into object entities by clustering SAM mask descriptors across views~\citep{cen2025laga}, while THGS constructs Gaussian
adjacency graphs with contrastive edge reweighting for hierarchical superpoint partitioning~\citep{dai2025training}. LUDVIG avoids training
altogether by combining multi-view weighted feature averaging with graph diffusion across adjacent primitives~\citep{marrie2025ludvig}. Although
these approaches eliminate dense feature optimization, they introduce multi-stage grouping heuristics and remain sensitive to primitive spatial overlap.

More recent methods avoid additional optimization by directly inverting the image-to-primitive aggregation used during rendering. Occam's LGS~\citep{cheng2024occam}, CosegGaussians~\citep{dou2024learning}, and DrSplat~\citep{jun2025dr} independently use row-normalized splatting equations as zero-shot feature lifter. %

\begin{wrapfigure}{r}{0.44\textwidth}
\centering
\resizebox{1.06\linewidth}{!}{%
  \input{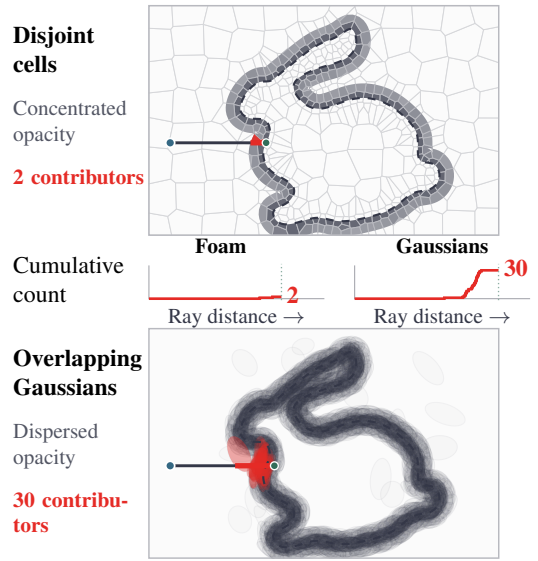}%
}
\caption{\textbf{Illustration of ray contributions.}
Disjoint cells with concentrated opacity and overlapping
Gaussians. Middle plots show cumulative
contributor counts.}
\label{fig:overview}
\end{wrapfigure}
Splat Feature Solver (SFS)~\citep{xiong2026sfs} formulates
feature lifting as a sparse linear inverse problem and provides a closed-form row-sum
estimator with a provable approximation bound. Specifically, SFS defines $\beta$ as a measure of the dispersion of the optimal lifted feature along a viewing ray.
In the case where each ray intersects only a single primitive, each row of the system matrix contains a single non-zero entry of value one, yielding this zero-dispersion regime.
Standard 3D Gaussians, however, have substantial spatial overlap, causing multiple primitives to contribute to the same ray and increasing feature dispersion (Figure~\ref{fig:overview}). SFS therefore introduces Tikhonov Guidance, which modifies the opacity weights during lifting to increase diagonal dominance and reduce this coupling. Thus, while SFS provides a principled formulation and efficient solver, its lifting accuracy remains tied to the overlap structure of the underlying Gaussian primitives.

An alternative to mitigating primitive overlap through the lifting procedure is to eliminate overlap at the representation level. Space-partitioning representations divide the scene into disjoint spatial cells, so each 3D location is associated with a single primitive by construction. Radiant Foam~\citep{govindarajan2025radiant} represents a scene using non-overlapping volumetric cells based on an unbounded Voronoi construction. However, its ray-tracing procedure relies on explicit cell stepping rather than tile-based rasterization, making it difficult to evaluate under the same rendering pipeline as standard 3D Gaussian Splatting. PowerFoam~\citep{govindarajan2026power} instead generalizes foam representations to bounded power diagrams, defining finite spatial cells through seed sites and per-cell radii. These bounded cells support tile-based rasterization while preserving disjoint spatial boundaries, making PowerFoam a suitable representation for studying feature lifting under otherwise comparable rendering conditions.

 Semantic Foam~\citep{sharafeldin2026semantic} jointly learns
reconstruction and object-identity features supervised by
segmentation masks. These features encode scene-specific
identities rather than text-aligned semantics, so they cannot
be directly compared with text embeddings for open-vocabulary
queries. In contrast, we study feature transfer from fixed 2D observations onto an already trained reconstruction, allowing the effect of the underlying representation on lifting to be isolated from semantic optimization. Building on this setting, we introduce \emph{OV-Foam}, an open-vocabulary 3D segmentation framework that lifts 2D vision-language features onto PowerFoam reconstructions.

\section{Method}
\label{sec:method}

OV-Foam lifts image features onto a PowerFoam reconstruction. We first describe how the cells define rendering contributions, then use those contributions in the SFS feature estimator. We then examine how image observations are distributed across primitives and whether each primitive receives consistent features from one class or conflicting features from several classes.

\subsection{PowerFoam}
PowerFoam represents a scene with a power diagram~\citep{govindarajan2026power}. Each primitive $j$ is associated with a center $\mathbf{p}_j$ and radius $a_j$. For a 3D point x, its power cell $V_j$ is the region where primitive $ j$ has the smallest power distance among all primitives $k$. The rendering support $C_j$ further restricts this cell to the primitive's bounding sphere:
\begin{equation}
\begin{aligned}
 V_j&=\{x:\|x-\mathbf{p}_j\|^2-a_j^2\leq\|x-\mathbf{p}_k\|^2-a_k^2\ \text{for all }k\},\\
 C_j&=V_j\cap\operatorname{Sphere}(\mathbf{p}_j,a_j).
\end{aligned}
\label{eq:cells}
\end{equation}
The power cells partition space, so their interiors do not overlap, although a ray can traverse several cells. Each primitive is bounded to $C_j$, so only the portion of its cell inside the bounding sphere can contribute to rendering.

Within each support $C_j$, a dipole defines an oriented boundary separating occupied space from empty space. The dipole provides a planar reference boundary, where occupied space satisfies $(\mathbf{x} - \mathbf{p}_j) \cdot \mathbf{n}_j < 0$, with $\mathbf{n}_j$ denoting the outward unit normal and $\mathbf{p}_j$ the primitive center. A displacement field can deform this boundary away from the reference plane. For a given ray $i$, PowerFoam determines the resulting intersections with both the support $C_j$ and the occupied region. These intersections define the occupied ray segment and its length $\ell_{ij}$. Density is constant within the occupied region and zero elsewhere. During reconstruction, primitive centers, radii, densities, dipoles, and enabled displacement parameters are optimized.

For a ray $r_i(t)$ parameterized by distance, volume rendering of a continuous 3D feature field $f: \mathbb{R}^3 \to \mathbb{R}^F$ between near and far limits $t_n$ and $t_f$ gives:
\begin{equation}
    \mathcal{F}_i = \int_{t_n}^{t_f} T_i(t) \sigma(r_i(t)) f(r_i(t)) \, dt, \qquad T_i(t) = \exp \!\left( -\int_{t_n}^{t} \sigma(r_i(s)) \, ds \right),
    \label{eq:foam-integral}
\end{equation}
where $\mathcal{F}_i$ is the rendered feature for ray $i$, and $T_i(t)$ is the transmittance accumulated before distance $t$. PowerFoam discretizes $f(\mathbf{x})$ as a piecewise-constant field across primitives, where $f(\mathbf{x}) = x_j \in \mathbb{R}^F$ within the occupied region of primitive $j$ and zero elsewhere. Because each primitive has constant density within its occupied region and carries a single feature $x_j$, the contribution of each occupied ray segment can be expressed through its length $\ell_{ij}$. For primitive $j$, a segment of length $\ell_{ij}$ and density $\sigma_j$ has opacity $\alpha_{ij} = 1 - \exp(-\sigma_j \ell_{ij})$. For the occupied segment of each primitive, the corresponding transmittance and opacity give the front-to-back weights:
\begin{equation}
    A_{ij} = T_{ij} \alpha_{ij}, \qquad T_{ij} = \prod_{k \prec_i j} (1 - \alpha_{ik}), \qquad \mathcal{F}_i = \sum_j A_{ij} x_j, \quad s_i = \sum_j A_{ij} \leq 1.
    \label{eq:weights}
\end{equation}
Here $k \prec_i j$ denotes a primitive encountered before $j$ along ray $i$, and the background feature is zero. The total scene contribution can equivalently be written as $s_i = 1 - \prod_j (1 - \alpha_{ij})$, with the remaining $1 - s_i$ reaching the background. Thus, $A_{ij}$ measures the contribution of primitive $j$ to pixel $i$, after absorption by earlier primitives. A shorter occupied segment reduces $\alpha_{ij}$, while opacity accumulated earlier along the ray reduces $T_{ij}$.

Gaussian splatting uses the same front-to-back compositing formulation, but obtains $\alpha_{ij}$ from the opacity and projected footprint of each Gaussian~\citep{kerbl2023gaussian}. PowerFoam instead integrates density through bounded occupied segments. Both renderers therefore provide weights $A_{ij}$ for feature lifting, despite their different geometry. Spatial partitioning prevents overlapping occupied interiors, but contributions from different primitives can still accumulate along the same ray, depending on the reconstructed surface and density.

\subsection{Feature lifting}
Let $B \in \mathbb{R}^{R \times F}$ contain $F$ feature values for each of the $R$ rays, and let $X \in \mathbb{R}^{P \times F}$ contain the unknown features of $P$ primitives. With the reconstruction fixed, the renderer defines a nonnegative contribution matrix $A \in \mathbb{R}^{R \times P}$, whose entries $A_{ij}$ describe how much primitive $j$ contributes to ray $i$. Rendering the primitive features $X$ with these contributions produces the image features $AX$, which can be compared to the observed features $B$. SFS therefore formulates feature recovery as a least-squares inverse problem~\citep{xiong2026sfs}:
\begin{equation}
    \widehat{X} =\operatorname*{arg\,min}_X \mathcal{L}(X), \qquad \mathcal{L}(X) = \frac{1}{2} \|AX - B\|^2.
    \label{eq:feature-ls}
\end{equation}
The minimum-norm least-squares is $A^\dagger B$, where $A^\dagger$ is the Moore--Penrose pseudoinverse. This solution minimizes the residual, but does not necessarily reproduce $B$ exactly. We use $\widehat X=A^\dagger B$ to define dispersion in Section~\ref{sec:dispersion}. The SFS one-shot approximation instead estimates each
primitive feature by averaging the observations it receives:
\begin{equation}
 x'_j=\frac{\sum_i A_{ij}B_i}{\sum_i A_{ij}},\qquad
 X'=D^{-1}A^\top B,\qquad D=\diag(A^\top\one).
 \label{eq:lift}
\end{equation}
This expression applies to primitives with positive total contribution. Here $D_{jj}=\sum_iA_{ij}$ is that total and $\one\in\R^R$ is the all-ones vector. The average minimizes the separable objective
\begin{equation}
 J(X)=\tfrac12\sum_{i,j}A_{ij}\|x_j-B_i\|_2^2.
 \label{eq:lift-surrogate}
\end{equation}

Unlike $\loss$, $J$ compares each primitive directly with each observation, rather than comparing their rendered mixture with that observation. SFS also gives a squared-weight variant after opacity adjustment in its Appendix~B, while Equation~\ref{eq:lift} corresponds to its base estimator. Within the OV-Foam lifting framework, we instantiate this estimator on PowerFoam reconstructions by accumulating the numerator and denominator using PowerFoam's rendering weights. Real image observations use SAM regions encoded with CLIP~\citep{kirillov2023sam,radford2021clip}.

For this estimator, the relationship to the least-squares solution depends on the number and strength of contributors along each ray. When every ray has one fully absorbing contributor, Equation~\ref{eq:lift} is an exact least-squares solution. With multiple contributors, the unknown features are generally coupled, making the independent average an approximation. Even with a single contributor, however, weak opacity does not guarantee equivalence to least squares, since incomplete absorption changes the feature scale required to reproduce the observation. Equation~\ref{eq:lift} normalizes by $\sum_i A_{ij}$, whereas the normal equations normalize by $\sum_i A_{ij}^2$; these agree only when every $A_{ij}\in\{0,1\}$.

\subsection{Dispersion and class purity}\label{sec:dispersion}
SFS introduces $\beta$ as a measure of feature dispersion along each viewing ray, characterizing how closely the one-shot estimator approximates the least-squares solution. Under complete absorption, ray opacity $s_i=1$. For each contributing primitive, define the feature discrepancy with respect to the observed pixel feature as $\Delta_{ij}=\|\widehat x_j-B_i\|_2$. Its weighted mean and relative dispersion are
\begin{equation}
 \mu_i=\sum_jA_{ij}\Delta_{ij},\qquad
 \beta_i=\frac{\sum_jA_{ij}(\Delta_{ij}-\mu_i)^2}{\mu_i^2},
 \qquad \mu_i>0.
 \label{eq:method-beta}
\end{equation}
The condition $s_i=1$ turns %
$A_{ij}$ into a probability distribution over  contributors along ray $i$. %
Low $\beta_i$ indicates that the primitives contributing to ray $i$ share similar error distances.

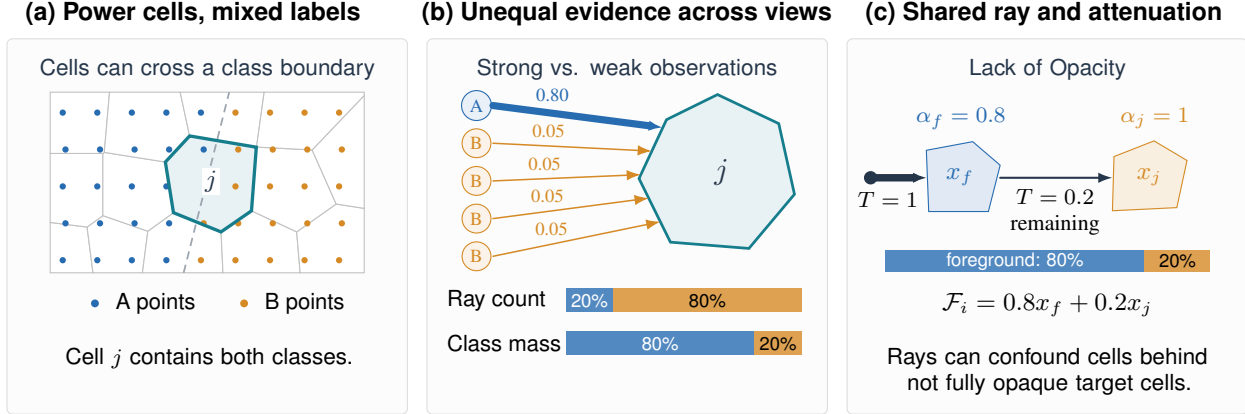
\begin{figure}[t]
\centering
\resizebox{\linewidth}{!}{%
\begin{tikzpicture}[x=1cm,y=1cm,every node/.style={},font=\sffamily\fontsize{9}{10.5}\selectfont,
  ldarrow/.style={-{Latex[length=1.8mm]},line cap=round},
  ldpanel/.style={rounded corners=2pt,draw=black!13,fill=black!1},
  ldsmall/.style={font=\sffamily\fontsize{8}{9.4}\selectfont,align=center,text width=4.6cm}]
\definecolor{ldA}{HTML}{276BB1}
\definecolor{ldB}{HTML}{D58A25}
\definecolor{ldMean}{HTML}{7551A4}
\definecolor{ldInk}{HTML}{25364A}
\definecolor{ldFoam}{HTML}{167D8D}

\path[use as bounding box] (0,.8) rectangle (15.8,6.15);
\foreach \x in {0,5.35,10.7}{\draw[ldpanel] (\x,.85) rectangle +(5.1,4.78);}
\node[anchor=west,font=\sffamily\bfseries\fontsize{9}{10.5}\selectfont] at (.1,5.95) {(a) Power cells, mixed labels};
\node[anchor=west,font=\sffamily\bfseries\fontsize{9}{10.5}\selectfont] at (5.15,5.95) {(b) Unequal evidence across views};
\node[anchor=west,font=\sffamily\bfseries\fontsize{9}{10.5}\selectfont] at (10.8,5.95) {(c) Shared ray and attenuation};
\node[ldsmall,text=ldInk] at (2.55,5.24) {Cells can cross a class boundary};
\path[fill=white,draw=black!24,line width=.4pt] (0.5500,2.6500)--(1.0952,2.6500)--(1.0200,3.2325)--(0.5500,3.2912)--cycle;
\path[fill=white,draw=black!24,line width=.4pt] (1.0952,2.6500)--(1.7253,2.6500)--(1.7938,3.1793)--(1.2215,3.3780)--(1.0200,3.2325)--cycle;
\path[fill=white,draw=black!24,line width=.4pt] (1.7253,2.6500)--(2.7510,2.6500)--(2.7262,3.1706)--(2.1203,3.4154)--(1.7938,3.1793)--cycle;
\path[fill=white,draw=black!24,line width=.4pt] (2.7510,2.6500)--(3.7337,2.6500)--(3.6899,3.1710)--(3.1141,3.4388)--(2.7262,3.1706)--cycle;
\path[fill=white,draw=black!24,line width=.4pt] (3.7337,2.6500)--(4.5500,2.6500)--(4.5500,3.4704)--(4.1928,3.5256)--(3.6899,3.1710)--cycle;
\path[fill=white,draw=black!24,line width=.4pt] (0.5500,4.1699)--(0.5500,3.2912)--(1.0200,3.2325)--(1.2215,3.3780)--(1.2215,4.1699)--cycle;
\path[fill=white,draw=black!24,line width=.4pt] (1.2215,4.1699)--(1.2215,3.3780)--(1.7938,3.1793)--(2.1203,3.4154)--(2.0171,4.0605)--(1.2394,4.1833)--cycle;
\path[fill=ldFoam!10,draw=black!24,line width=.4pt] (2.3260,4.3928)--(2.0171,4.0605)--(2.1203,3.4154)--(2.7262,3.1706)--(3.1141,3.4388)--(3.1758,4.2497)--cycle;
\path[fill=white,draw=black!24,line width=.4pt] (3.1758,4.2497)--(3.1141,3.4388)--(3.6899,3.1710)--(4.1928,3.5256)--(3.8866,4.2146)--(3.1781,4.2515)--cycle;
\path[fill=white,draw=black!24,line width=.4pt] (4.5500,3.4704)--(4.5500,4.9091)--(3.8866,4.2146)--(4.1928,3.5256)--cycle;
\path[fill=white,draw=black!24,line width=.4pt] (1.3061,4.9500)--(0.5500,4.9500)--(0.5500,4.1699)--(1.2215,4.1699)--(1.2394,4.1833)--cycle;
\path[fill=white,draw=black!24,line width=.4pt] (2.2058,4.9500)--(1.3061,4.9500)--(1.2394,4.1833)--(2.0171,4.0605)--(2.3260,4.3928)--cycle;
\path[fill=white,draw=black!24,line width=.4pt] (3.2757,4.9500)--(2.2058,4.9500)--(2.3260,4.3928)--(3.1758,4.2497)--(3.1781,4.2515)--cycle;
\path[fill=white,draw=black!24,line width=.4pt] (4.5500,4.9091)--(4.5500,4.9500)--(3.2757,4.9500)--(3.1781,4.2515)--(3.8866,4.2146)--cycle;
\draw[ldInk!45,densely dashed,line width=.65pt] (2.25,2.65)--(2.83,4.95);
\fill[ldA] (0.7100,2.8100) circle (.040);
\fill[ldA] (1.1500,2.8100) circle (.040);
\fill[ldA] (1.5900,2.8100) circle (.040);
\fill[ldA] (2.0300,2.8100) circle (.040);
\fill[ldB] (2.4700,2.8100) circle (.040);
\fill[ldB] (2.9100,2.8100) circle (.040);
\fill[ldB] (3.3500,2.8100) circle (.040);
\fill[ldB] (3.7900,2.8100) circle (.040);
\fill[ldB] (4.2300,2.8100) circle (.040);
\fill[ldA] (0.7450,3.2800) circle (.040);
\fill[ldA] (1.1850,3.2800) circle (.040);
\fill[ldA] (1.6250,3.2800) circle (.040);
\fill[ldA] (2.0650,3.2800) circle (.040);
\fill[ldB] (2.5050,3.2800) circle (.040);
\fill[ldB] (2.9450,3.2800) circle (.040);
\fill[ldB] (3.3850,3.2800) circle (.040);
\fill[ldB] (3.8250,3.2800) circle (.040);
\fill[ldB] (4.2650,3.2800) circle (.040);
\fill[ldA] (0.7100,3.7500) circle (.040);
\fill[ldA] (1.1500,3.7500) circle (.040);
\fill[ldA] (1.5900,3.7500) circle (.040);
\fill[ldA] (2.0300,3.7500) circle (.040);
\fill[ldB] (2.9100,3.7500) circle (.040);
\fill[ldB] (3.3500,3.7500) circle (.040);
\fill[ldB] (3.7900,3.7500) circle (.040);
\fill[ldB] (4.2300,3.7500) circle (.040);
\fill[ldA] (0.7450,4.2200) circle (.040);
\fill[ldA] (1.1850,4.2200) circle (.040);
\fill[ldA] (1.6250,4.2200) circle (.040);
\fill[ldA] (2.0650,4.2200) circle (.040);
\fill[ldA] (2.5050,4.2200) circle (.040);
\fill[ldB] (2.9450,4.2200) circle (.040);
\fill[ldB] (3.3850,4.2200) circle (.040);
\fill[ldB] (3.8250,4.2200) circle (.040);
\fill[ldB] (4.2650,4.2200) circle (.040);
\fill[ldA] (0.7100,4.6900) circle (.040);
\fill[ldA] (1.1500,4.6900) circle (.040);
\fill[ldA] (1.5900,4.6900) circle (.040);
\fill[ldA] (2.0300,4.6900) circle (.040);
\fill[ldA] (2.4700,4.6900) circle (.040);
\fill[ldB] (2.9100,4.6900) circle (.040);
\fill[ldB] (3.3500,4.6900) circle (.040);
\fill[ldB] (3.7900,4.6900) circle (.040);
\fill[ldB] (4.2300,4.6900) circle (.040);
\draw[ldFoam,line width=1.2pt] (2.3260,4.3928)--(2.0171,4.0605)--(2.1203,3.4154)--(2.7262,3.1706)--(3.1141,3.4388)--(3.1758,4.2497)--cycle;

\node[font=\fontsize{9}{10.5}\selectfont,fill=white,inner sep=1pt,text=ldInk] at (2.60,3.81) {$j$};
\fill[ldA] (1.12,2.26) circle (.05);
\node[anchor=west,font=\sffamily\fontsize{8}{9.4}\selectfont] at (1.25,2.26) {A points};
\fill[ldB] (3.02,2.26) circle (.05);
\node[anchor=west,font=\sffamily\fontsize{8}{9.4}\selectfont] at (3.15,2.26) {B points};
\node[ldsmall,text width=4.5cm] at (2.55,1.56)
 {Cell $j$ contains both classes.};

\node[ldsmall,text=ldInk] at (7.9,5.24)
  {Strong vs. weak observations};

\path[draw=ldFoam,fill=ldFoam!10,line width=1pt]
  (8.40,4.60)--
  (8.05,3.85)--
  (8.45,3.04)--
  (9.45,2.96)--
  (10.03,3.65)--
  (9.83,4.55)--
  (9.05,4.92)--cycle;

\node[text=ldInk,font=\fontsize{11}{12}\selectfont]
  at (9.10,3.94) {$j$};

\foreach \y/\ex/\ey/\c/\lab/\w/\wt in {
  4.78/8.3533/4.50/ldA/A/2.6/0.80,
  4.30/8.2133/4.20/ldB/B/.55/0.05,
  3.82/8.0733/3.90/ldB/B/.55/0.05,
  3.34/8.1735/3.60/ldB/B/.55/0.05,
  2.86/8.3216/3.30/ldB/B/.55/0.05
}{
  \node[
    circle,draw=\c,fill=\c!10,
    inner sep=1.5pt,
    font=\fontsize{6.7}{7.7}\selectfont,
    text=\c
  ] at (5.98,\y) {\lab};

  \draw[ldarrow,draw=\c,line width=\w pt]
    (6.19,\y) --
    node[
      pos=.35,above=3pt,
      inner sep=0pt,
      font=\fontsize{6.7}{7.7}\selectfont,
      text=\c
    ] {\wt}
    (\ex,\ey);
}

\node[anchor=west,font=\sffamily\fontsize{8}{9.4}\selectfont] at (5.47,2.30) {Ray count};
\fill[ldA!80] (7.12,2.15) rectangle (7.72,2.45);
\fill[ldB!80] (7.72,2.15) rectangle (10.12,2.45);
\node[font=\sffamily\fontsize{6.7}{7.7}\selectfont,text=white] at (7.42,2.30) {20\%};
\node[font=\sffamily\fontsize{6.7}{7.7}\selectfont] at (8.92,2.30) {80\%};
\node[anchor=west,font=\sffamily\fontsize{8}{9.4}\selectfont] at (5.47,1.75) {Class mass};
\fill[ldA!80] (7.12,1.60) rectangle (9.52,1.90);
\fill[ldB!80] (9.52,1.60) rectangle (10.12,1.90);
\node[font=\sffamily\fontsize{6.7}{7.7}\selectfont,text=white] at (8.32,1.75) {80\%};
\node[font=\sffamily\fontsize{6.7}{7.7}\selectfont] at (9.82,1.75) {20\%};

\node[ldsmall,text=ldInk] at (13.25,5.24) {Lack of Opacity};
\node[font=\fontsize{8}{9.4}\selectfont,text=ldA] at (12.15,4.62) {$\alpha_f=0.8$};
\node[font=\fontsize{8}{9.4}\selectfont,text=ldB] at (14.62,4.62) {$\alpha_j=1$};
\draw[ldarrow,draw=ldInk,line width=2.7pt] (11.04,3.86)--(11.72,3.86);
\path[draw=ldA,fill=ldA!14] (11.74,3.43)--(12.50,3.43)--(12.65,4.02)--(12.3,4.32)--(11.71,4.21)--cycle;
\draw[ldarrow,draw=ldInk,line width=.75pt] (12.65,3.86)--(14.08,3.86);
\path[draw=ldB,fill=ldB!14] (14.08,3.44)--(14.9,3.48)--(15.03,4.08)--(14.61,4.33)--(14.10,4.13)--cycle;
\node[text=ldA] at (12.14,3.87) {$x_f$};
\node[text=ldB] at (14.55,3.87) {$x_j$};
\fill[ldInk] (11.0,3.86) circle (.085);
\node[font=\fontsize{8}{9.4}\selectfont] at (11.22,3.58) {$T=1$};
\node[font=\fontsize{8}{9.4}\selectfont,align=center] at (13.35,3.43) {$T=0.2$\\remaining};
\fill[ldA!80] (11.18,2.66) rectangle (14.49,2.94);
\fill[ldB!80] (14.49,2.66) rectangle (15.32,2.94);
\node[font=\sffamily\fontsize{6.7}{7.7}\selectfont,text=white] at (12.84,2.80) {foreground: 80\%};
\node[font=\sffamily\fontsize{6.7}{7.7}\selectfont] at (14.91,2.80) {20\%};
\node[font=\fontsize{9}{10.5}\selectfont] at (13.25,2.24) {$\mathcal F_i=0.8x_f+0.2x_j$};
\node[ldsmall,text width=4.65cm] at (13.25,1.40)
 {Rays can confound cells behind\\not fully opaque target cells.};

\end{tikzpicture}
}
\vspace{-6mm}
\caption{\textbf{How semantic evidence reaches a primitive.}
(a) A cell can contain multiple classes.
(b) Observation weights can reverse the majority suggested by ray counts.
(c) Partial transparency lets foreground observations influence cells behind the surface.}
\label{fig:mixing}
\end{figure}

In addition, we introduce purity $P_j$, which addresses the complementary question of which classes contribute evidence to a primitive. Given labeled image rays with classes $y_i$, let
\begin{equation}
 \pi_{jc}=\frac{\sum_{i:y_i=c}A_{ij}}{\sum_{i:y_i\text{ is labeled}}A_{ij}},
 \qquad P_j=\max_c\pi_{jc}.
 \label{eq:method-purity}
\end{equation}
Where $\pi_{jc}$ is the share of the rendering \emph{weight} that primitive $j$ receives from class $c$, not the share of rays: one ray striking $j$ with a large $A_{ij}$ outweighs many that graze it, so the dominant class by weight can differ from the dominant class by count (Figure~\ref{fig:mixing}). Let $t_c$ be the unit CLIP text embedding of class $c$. With $B_i=t_{y_i}$ and the same labeled rays in both sums of Equation~\ref{eq:lift}, the lifted feature becomes $x'j=\sum_c\pi{jc}t_c$, a convex combination of class embeddings in which $P_j$ is the weight on the dominant class.

Inconsistent image observations arise from two distinct sources: geometric mixing, where rays sweep through multiple primitives across object boundaries, and upstream feature noise, where 2D models like CLIP produce different features across different viewpoints. These inconsistencies reveal a fundamental limitation of feature lifting: because each primitive $j$ stores a single feature vector $x_j \in \mathbb{R}^F$, any primitive receiving conflicting observations across views is forced to compromise. Even the exact least-squares solution $\widehat{X} = A^\dagger B$ returns a blended feature vector (e.g., a centroid between competing class embeddings). While this centroid minimizes the quadratic rendering loss from \autoref{eq:feature-ls}, its diluted feature vector fails to align cleanly with any single class embedding during text-prompt classification.

We therefore distinguish the solver approximation error from the unavoidable semantic error of the least-squares target. By the orthogonality of least-squares residuals ($A^\top(A\widehat{X} - B) = 0$), the additional reconstruction error incurred by the one-shot estimator $X'$ is defined by the \emph{fitting gap}:
\begin{equation}
\mathcal{L}(X') - \mathcal{L}(\widehat{X}) = \frac{1}{2} \|A(X' - \widehat{X})\|^2 \le \frac{1}{2} \sum_{i=1}^R \sum_{j,k} A_{ij} A_{ik} \|\widehat{x}_j - \widehat{x}_k\|_2^2.
\label{eq:fitting-gap}
\end{equation}
The inequality additionally assumes complete absorption, $s_i=1$, the same row-stochastic condition assumed by SFS's own approximation bound~\citep{xiong2026sfs}. Here, $A_{ij} A_{ik}$ weights the co-contribution of primitives $j$ and $k$ along ray $i$, while $\|\widehat{x}_j - \widehat{x}_k\|_2^2$ captures their pairwise feature disagreement. This inequality shows that the fitting gap is bounded by the shared-ray feature disagreement between co-contributing primitives. Ray dispersion $\beta$ (Eq.~\ref{eq:method-beta}) provides a complementary ray-level measure of this contributor disagreement.
Concentrating ray weights onto fewer primitives reduces these cross-contributor terms, hence reducing the upper bound on the fitting gap $\mathcal{L}(X') - \mathcal{L}(\widehat{X})$ and in extension allowing for better feature assignments.

However, because a small fitting gap cannot fix semantic ambiguity in the least-squares optimum $\widehat{X}$, we evaluate primitive purity $P_j$ to measure whether the evidence aggregated at primitive $j$ is dominated by a single class. Together, $\beta$, the fitting gap, and purity allow us to trace how the 3D reconstruction governs feature lifting: from how closely $X'$ approximates $\widehat{X}$, to the class consistency of the evidence being aggregated, and ultimately to the accuracy of the recovered 3D labels.
Figure~\ref{fig:mixing} illustrates how spatial ownership, observation weights, and shared rays govern this interplay.

To disentangle these effects, we separate reconstruction-dependent errors from errors in the semantic observations serving as input to the lifting. We first isolate the geometric quality of the reconstruction by lifting ground-truth class masks encoded as text embeddings. Because both reconstructions receive perfect upstream labels, any difference in the resulting 3D labels proceeds from reconstruction and lifting process, %
unaffected by CLIP noise. We then evaluate open-vocabulary segmentation using real SAM/CLIP image features, testing whether the reconstruction-dependent effects persist under practical, noisy observation conditions.

\section{Experiments}
\label{sec:experiments}
We evaluate the effect of the reconstruction on feature lifting at three levels. First, we compare the dispersion $\beta$ and the fitting gap $\mathcal{L}(X') - \mathcal{L}(\widehat{X})$ to determine whether the observed differences in $\beta$ correspond to differences in the quality of the one-shot lifting solution. We then examine purity and primitive-level prediction accuracy to determine whether the evidence accumulated by each reconstruction is more class-consistent.

\enlargethispage{2\baselineskip}
To isolate the effect of the reconstruction from errors in the input image features, we next lift ground-truth image labels encoded as class text embeddings. This controlled experiment gives both reconstructions the same correct semantic observations and therefore tests whether differences in the recovered 3D labels arise from the reconstruction and lifting process itself. Finally, we evaluate CLIP image features, measuring whether the same differences persist when the lifted observations also contain errors.

\subsection{Dispersion and class purity}
\label{sec:beta}\label{sec:evidence}

We compare PowerFoam and 3DGS on ten ScanNet
scenes~\citep{dai2017scannet}, initializing both from the
same GT vertices with one primitive per vertex, following the OpenGaussian~\citep{wu2024opengaussian} protocol. Centers
and primitive counts remain fixed during reconstruction,
and all reconstruction parameters remain fixed during
lifting. We measure the number of nonzero contributors
per ray, $k_i$, and dispersion $\beta_i$, using a least-squares reference computed on twelve
matched views per scene. PSNR uses all views at native resolution.

\begin{wraptable}{r}{0.46\linewidth}
\vspace{-7.75mm}
\centering
\caption{\textbf{Dispersion and ray contribution.}
Ten-scene averages. Dispersion is the mean of scene medians.
Per-scene results in Table~\ref{tab:matched-full}.}
\label{tab:matched}
\small
\setlength{\tabcolsep}{3pt}
\vspace{2.5mm}
\begin{tabular}{lrrr}
\toprule
Representation & PSNR $\uparrow$ & $k_i$ $\downarrow$
& $\beta_i$ $\downarrow$ \\
\midrule
PowerFoam & \textbf{27.01} & \textbf{2.32}
& \textbf{0.0080} \\
3DGS & 25.22 & 33.63 & 1.5554 \\
\bottomrule
\end{tabular}
\vspace{-2mm}
\end{wraptable}

Table~\ref{tab:matched} shows that PowerFoam uses fewer contributors per ray $k_i$ and exhibits lower dispersion $\beta_i$ while retaining
higher average RGB reconstruction quality. To establish
whether its one-shot estimate lies closer to least
squares, we measure the fitting gap using GT image labels
encoded as unit class text embeddings on twelve matched
views. We compare raw, unnormalized feature fields against
numerically converged, unregularized least-squares references.
Averaging the per-ray gap
$[\mathcal{L}(X')-\mathcal{L}(\widehat{X})]/R$
equally across scenes gives $4.663\times10^{-3}$ for 3DGS
and $8.352\times10^{-4}$ for PowerFoam, an $82.1\%$
reduction. PowerFoam has the smaller gap and lower one-shot
loss in all ten scenes, although 3DGS reaches the lower
least-squares reference loss. Thus, PowerFoam supports
a more accurate one-shot approximation on these
reconstructions, rather than a lower attainable
least-squares loss.

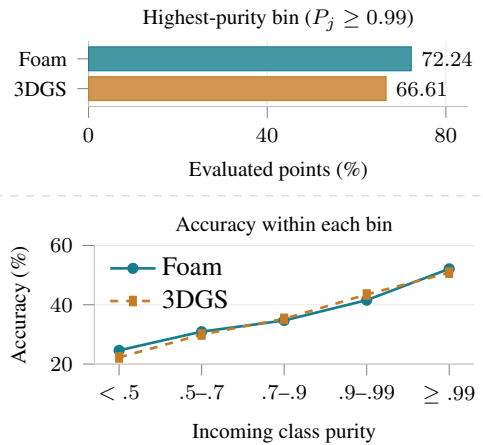
\begin{wrapfigure}{r}{.4\textwidth}
\vspace{-7mm}
\centering
\definecolor{foam}{HTML}{167D8D}
\definecolor{gs}{HTML}{C67B27}

\begin{tikzpicture}
\begin{axis}[
 width=\linewidth,height=2.45cm,
 xbar,bar width=9pt,bar shift=0pt,
 xmin=0,xmax=85,
 ymin=-.6,ymax=1.6,
 ytick={0,1},
 yticklabels={3DGS,Foam},
 xtick={0,40,80},
 xlabel={Evaluated points (\%)},
 title={Highest-purity bin ($P_j\geq0.99$)},
title style={font=\scriptsize,yshift=-6pt},
 tick label style={font=\scriptsize},
 label style={font=\scriptsize},
 axis lines*=left,
 tick align=outside,
 axis line style={gray!50},
 xmajorgrids,grid style={gray!15},
 nodes near coords,
 nodes near coords style={
   font=\scriptsize,
   /pgf/number format/fixed,
   /pgf/number format/precision=2,
   /pgf/number format/fixed zerofill
 }
]
\addplot[fill=foam!80,draw=foam]
 coordinates {(72.24,1)};
\addplot[fill=gs!80,draw=gs]
 coordinates {(66.61,0)};
\end{axis}
\end{tikzpicture}

\par\vspace{-4mm}
\noindent
\begin{tikzpicture}
\draw[gray!35,dashed,line width=.4pt]
  (0,0) -- (\linewidth,0);
\end{tikzpicture}
\par\vspace{1mm}

\begin{tikzpicture}
\begin{axis}[
 width=\linewidth,height=3.15cm,
 xmin=.7,xmax=5.3,
 ymin=20,ymax=60,
 xtick={1,2,3,4,5},
 xticklabels={
   $<.5$,
   $.5$--$.7$,
   $.7$--$.9$,
   $.9$--$.99$,
   $\geq.99$
 },
 ytick={20,40,60},
 xlabel={Incoming class purity},
 ylabel={Accuracy (\%)},
 title={Accuracy within each bin},
 title style={font=\scriptsize,yshift=-6pt},
 tick label style={font=\scriptsize},
 label style={font=\scriptsize},
 axis lines*=left,
 tick align=outside,
 axis line style={gray!50},
 ymajorgrids,grid style={gray!15},
 legend style={
   font=\scriptsize,
   at={(.03,.97)},anchor=north west,
   draw=none,fill=none,
   inner sep=1pt
 },
 legend cell align=left
]
\addplot[
 color=foam,line width=1pt,
 mark=*,mark size=1.6pt
] coordinates {
 (1,24.63) (2,30.93) (3,34.70)
 (4,41.56) (5,52.13)
};
\addlegendentry{Foam}

\addplot[
 color=gs,line width=1pt,dashed,
 mark=square*,mark size=1.5pt
] coordinates {
 (1,22.24) (2,29.96) (3,35.27)
 (4,43.47) (5,50.83)
};
\addlegendentry{3DGS}
\end{axis}
\end{tikzpicture}
\captionsetup{font=footnotesize,skip=2pt}
\caption{\textbf{CLIP feature purity.}
Foam has more high-purity points.
Higher purity tracks higher accuracy.}
\label{fig:purity-real}
\vspace{-7.8mm}
\end{wrapfigure}

Closer agreement with least squares does not establish
whether a primitive receives consistent class evidence.
We therefore measure purity from
the evidence accumulated by the renderer and compare it
with prediction accuracy when lifting CLIP features.
Each observation contributes according to its rendering
weight $A_{ij}$, and purity measures the fraction of
accumulated class evidence belonging to the dominant class.

Figure~\ref{fig:purity-real} shows that accuracy increases
with purity across all five bands for both representations.
PowerFoam places more evaluated points in the highest-purity
band, where predictions are most accurate. Its advantage
therefore includes a more favorable distribution of
incoming evidence, rather than consistently higher
accuracy within every band. Purity tracks prediction
accuracy even with real image features, but does not
guarantee correct labels.

Even points assigned to the highest-purity primitives
remain frequently misclassified with real CLIP features.
Their contributing rays mostly observe the same GT class,
but the image features need not identify that class
correctly. We therefore replace the image features with
GT class embeddings to measure the segmentation errors
that remain when the upstream labels are correct.

\subsection{Lifting ground-truth semantics}
\label{sec:oracle}

To separate errors in the image features from errors
introduced by reconstruction and lifting, we replace each
labeled pixel's feature with the unit CLIP text embedding
of its GT class, obtained from ScanNet's label masks. Unlabeled pixels contribute no evidence.
With each reconstruction fixed, we apply
Equation~\ref{eq:lift} over all available lifting views
and classify the resulting primitive features by cosine
similarity to the class text embeddings.

\begin{wraptable}[9]{r}{.42069\textwidth}
\vspace{-2.6pt}
\centering
\small
\setlength{\tabcolsep}{6pt}
\caption{\textbf{Lifting ground-truth image semantics.}
The geometric upper bound (Geom UB) assigns each primitive the
majority GT class of its evaluated points.}
\label{tab:oracle}
\vspace{-1.5mm}
\resizebox{\linewidth}{!}{
\begin{tabular}{lccc}
\toprule
& \multicolumn{2}{c}{GT feature lifting}
& Geom UB \\
\cmidrule(lr){2-3}\cmidrule(lr){4-4}
Representation & Point accuracy & mIoU & mIoU \\
\midrule
PowerFoam & \textbf{96.37} & \textbf{90.59} & 98.13 \\
3DGS & 94.25 & 85.40 & \textbf{98.92} \\
\bottomrule
\end{tabular}}
\vspace{-2mm}
\end{wraptable}

Table~\ref{tab:oracle} shows that PowerFoam recovers more
accurate point labels from the same correct image
observations, with higher mIoU in all ten scenes.
This advantage is not explained by a higher
geometric upper bound. The geometric upper bound pinpoints the limit introduced by mixed labels, as described in Figure~\ref{fig:mixing}. Therefore, we conclude that PowerFoam
recovers more of the label accuracy available because of improved assignment from the image-based lifting.

These results complement the fitting-gap measurement.
PowerFoam's one-shot estimate is closer to its
least-squares reference, and, when supplied with correct
image labels, it also produces more accurate 3D
segmentation. The remaining question is whether this
advantage persists with imperfect image features and
different aggregation methods. We evaluate that setting
next by replacing the Gaussian representation in
existing open-vocabulary segmentation methods with
PowerFoam.

\subsection{Open-vocabulary segmentation}
\label{sec:benchmark}
We adapt existing Gaussian-based methods' feature lifting approaches to PowerFoam. Each method uses the same SAM/CLIP image features
for both representations, allowing us to compare
their segmentation results while holding the
upstream observations fixed.

For ScanNet, we follow OpenGaussian's ten-scene point-level
evaluation setup~\citep{wu2024opengaussian}, with the changes
to point assignment and opacity filtering described below.
We assign each evaluated point its primitive's predicted
class and report mIoU (mean intersection over union) and mAcc (mean class accuracy).
Predictions use three different class sets.
For PowerFoam, we assign points using the reconstructed
surface inside each cell, represented by its dipole,
rather than the cell center alone.
This surface can shift during reconstruction even when
the center remains fixed, so the center need not lie
in the primitive's occupied region
(Appendix~\ref{app:dipole-assignment}).

OpenGaussian's protocol excludes ground-truth points associated with
Gaussians whose opacity is below $0.1$.
However, low-opacity primitives can still contribute to
rendering and receive features during lifting.
Excluding their associated points removes part of the
reconstruction from assessment and allows the evaluated
point set to depend on the representation. It also means that a higher score on retained points need not imply better performance on the full scene.
We therefore retain all valid labeled ground-truth points
for both representations, so weakly represented regions
remain part of the segmentation task.
Under this common point set, PowerFoam improves both
metrics for every tested method across all three
class sets (Table~\ref{tab:scannet-unfiltered}). For the opacity-filtered evaluation, see Table \ref{tab:scannet-main}.

\begin{table}[h]
\centering
\setlength{\tabcolsep}{3pt}
\caption{\textbf{Open-vocabulary 3D segmentation on ScanNet.}
We report mIoU and mAcc (\%) for three class sets.
Bold marks the higher score within each pair.
Green parentheses show OV-Foam's improvement over 3DGS
in percentage points.
Appendix~\ref{app:scannet-eval} details the protocol.}
\label{tab:scannet-unfiltered}
\vspace{-3mm}
\begingroup
\small %
\newcommand{\deltaval}[1]{\textcolor{green!50!black}{+#1}}
\resizebox{\textwidth}{!}{
\begin{tabular}{llrrrrrr}
\toprule
& & \multicolumn{2}{c}{19 classes}
  & \multicolumn{2}{c}{15 classes}
  & \multicolumn{2}{c}{10 classes} \\
\cmidrule(lr){3-4}
\cmidrule(lr){5-6}
\cmidrule(lr){7-8}
Method & Representation
& mIoU & mAcc & mIoU & mAcc & mIoU & mAcc \\
\midrule
VALA~\citep{wang2025vala}
& 3DGS
& 30.08 & 47.57 & 32.70 & 52.00 & 39.56 & 60.35 \\
& OV-Foam
& \textbf{33.68} & \textbf{51.18} & \textbf{37.15} & \textbf{56.08} & \textbf{45.68} & \textbf{65.39} \\
&  $\Delta$
& \deltaval{3.60} & \deltaval{3.61} & \deltaval{4.45} & \deltaval{4.08} & \deltaval{6.12} & \deltaval{5.04} \\
\addlinespace[3pt]

Occam's LGS~\citep{cheng2024occam}
& 3DGS
& 32.76 & 50.40 & 35.62 & 55.24 & 43.94 & 64.66 \\
& OV-Foam
& \textbf{34.19} & \textbf{52.34} & \textbf{37.58} & \textbf{57.46} & \textbf{46.46} & \textbf{66.64} \\
& $\Delta$
& \deltaval{1.43} & \deltaval{1.94} & \deltaval{1.96} & \deltaval{2.22} & \deltaval{2.52} & \deltaval{1.98} \\
\addlinespace[3pt]

SFS~\citep{xiong2026sfs}
& 3DGS
& 32.03 & 49.29 & 35.41 & 54.78 & 44.34 & 64.24 \\
& OV-Foam
& \textbf{34.00} & \textbf{51.75} & \textbf{37.59} & \textbf{57.43} & \textbf{46.91} & \textbf{66.64} \\
& $\Delta$
& \deltaval{1.97} & \deltaval{2.46} & \deltaval{2.18} & \deltaval{2.65} & \deltaval{2.57} & \deltaval{2.40} \\
\addlinespace[3pt]

LUDVIG~\citep{marrie2025ludvig}
& 3DGS
& 32.94 & 50.36 & 36.58 & 56.22 & 45.64 & 65.43 \\
& OV-Foam
& \textbf{34.58} & \textbf{52.14} & \textbf{38.11} & \textbf{57.84} & \textbf{47.59} & \textbf{67.04} \\
& $\Delta$
& \deltaval{1.64} & \deltaval{1.78} & \deltaval{1.53} & \deltaval{1.62} & \deltaval{1.95} & \deltaval{1.61} \\
\bottomrule
\end{tabular}}
\endgroup
\end{table}

\begin{table}[t]
\centering\small\setlength{\tabcolsep}{7pt}
\caption{\textbf{LERF text-query evaluation.}
Mean IoU and mean accuracy across four scenes under a query and mask protocol. Bold marks the better representation.
$\Delta$ denotes OV-Foam minus 3DGS.}
\label{tab:lerf-main}
\begin{tabular}{lrrrrrr}
\toprule
 & \multicolumn{3}{c}{mIoU} & \multicolumn{3}{c}{mAcc}\\
\cmidrule(lr){2-4}\cmidrule(lr){5-7}
Method & 3DGS & OV-Foam  & $\Delta$
       & 3DGS & OV-Foam  & $\Delta$\\
\midrule
VALA~\citep{wang2025vala}
& 0.555 & \textbf{0.570} & \textcolor{green!50!black}{+0.015}
& 0.707 & \textbf{0.730} & \textcolor{green!50!black}{+0.023}\\
Occam's LGS~\citep{cheng2024occam}
& 0.522 & \textbf{0.529} & \textcolor{green!50!black}{+0.007}
& 0.699 & \textbf{0.749} & \textcolor{green!50!black}{+0.050}\\
SFS~\citep{xiong2026sfs}
& 0.502 & \textbf{0.570} & \textcolor{green!50!black}{+0.068}
& 0.641 & \textbf{0.659} & \textcolor{green!50!black}{+0.018}\\
LUDVIG~\citep{marrie2025ludvig}
& 0.438 & \textbf{0.543} & \textcolor{green!50!black}{+0.105}
& 0.735 & \textbf{0.746} & \textcolor{green!50!black}{+0.011}\\
\bottomrule
\end{tabular}
\end{table}

\begin{wrapfigure}{r}{.46\textwidth}
\vspace{-6.5mm}
\centering
\begingroup
\setlength{\tabcolsep}{0pt}

\begin{tabular}{
  @{}c@{\hspace{2pt}}
  c@{\hspace{1pt}}
  c@{\hspace{1pt}}c@{}
}
& {\scriptsize Cube}
& {\scriptsize Elephant}
& {\scriptsize Duck}\\[2pt]

\rotatebox{90}{\scriptsize SFS Foam}
&
\raisebox{-.5\height}{\includegraphics[width=.29\linewidth,trim={430bp 0bp 236bp 408bp},clip]{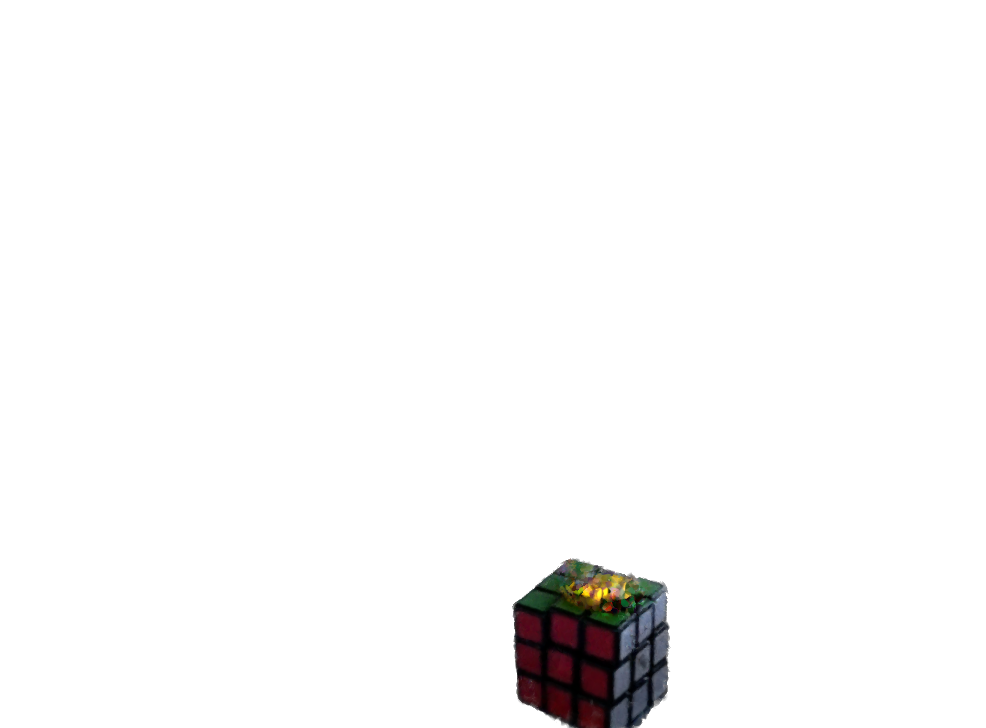}}%
&
\raisebox{-.5\height}{\includegraphics[width=.29\linewidth,trim={0bp 128bp 666bp 280bp},clip]{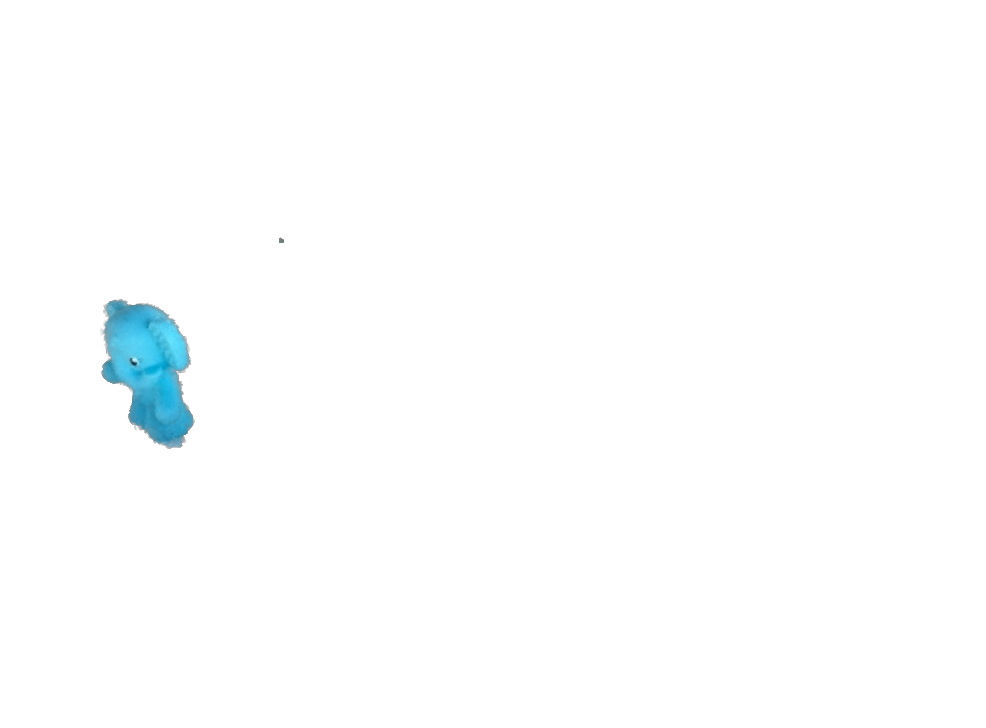}}%
&
\raisebox{-.5\height}{\includegraphics[width=.29\linewidth,trim={440bp 0bp 226bp 408bp},clip]{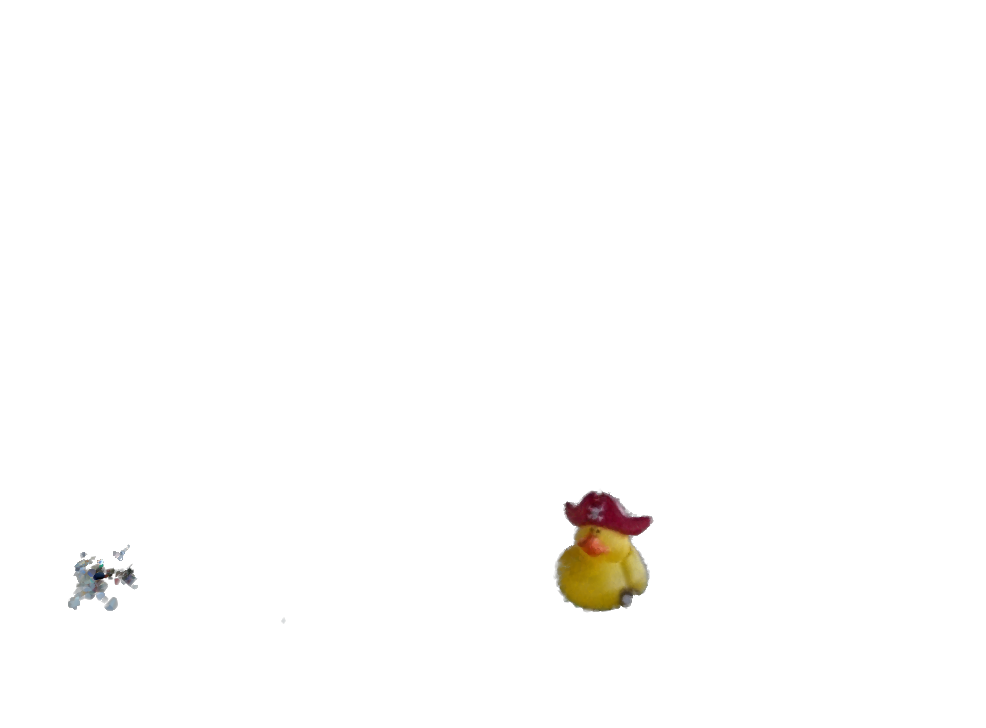}}%
\\[2pt]

\rotatebox{90}{\scriptsize SFS 3DGS}
&
\raisebox{-.5\height}{\includegraphics[width=.29\linewidth,trim={430bp 0bp 236bp 408bp},clip]{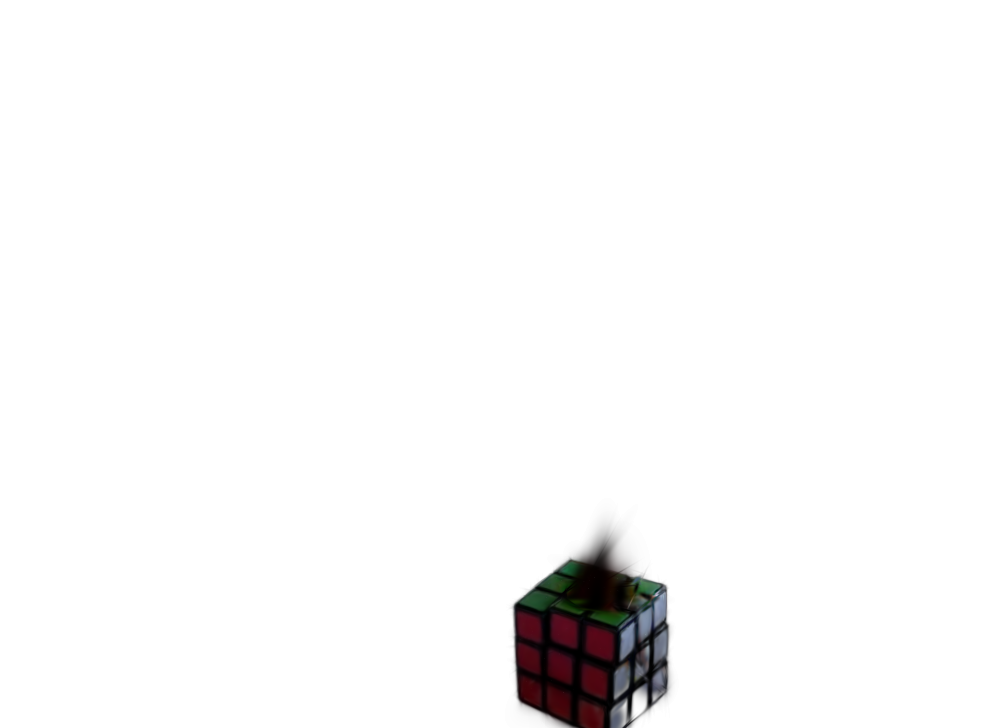}}%
&
\raisebox{-.5\height}{\includegraphics[width=.29\linewidth,trim={0bp 128bp 666bp 280bp},clip]{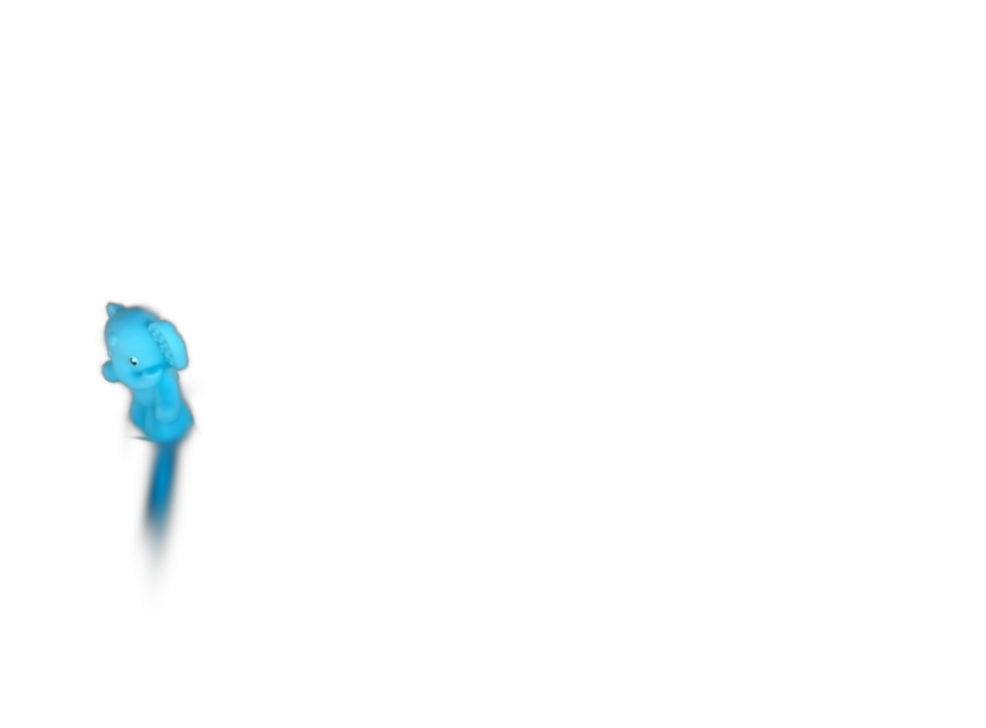}}%
&
\raisebox{-.5\height}{\includegraphics[width=.29\linewidth,trim={440bp 0bp 226bp 408bp},clip]{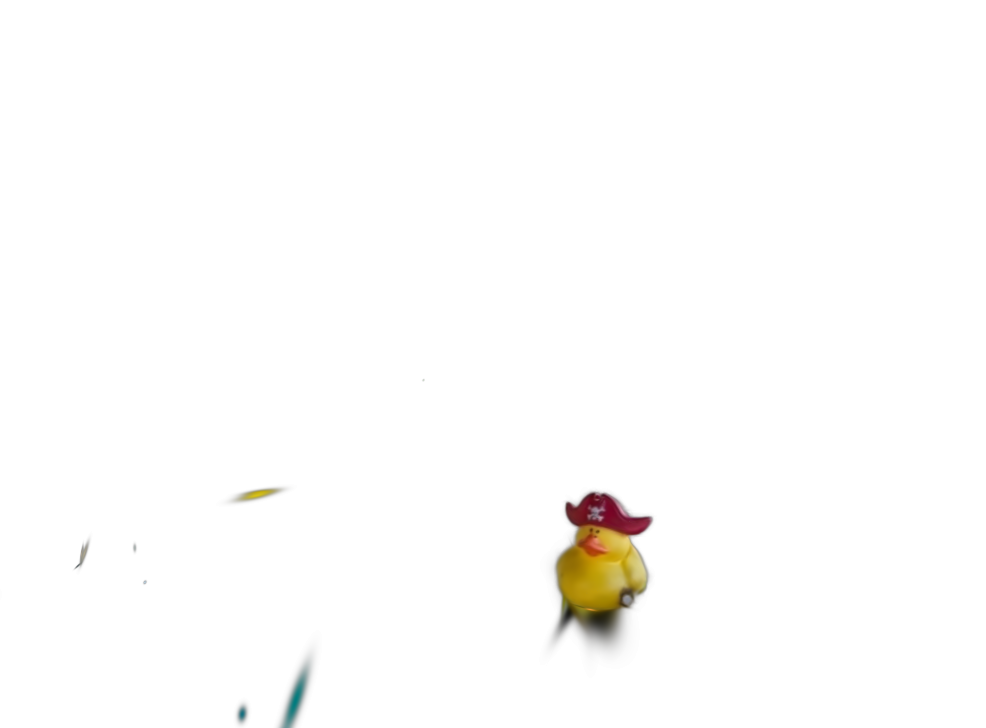}}%
\\
\end{tabular}
\endgroup
\label{fig:figurines-comparison}
\vspace{-5mm}
\end{wrapfigure}

We also evaluate rendered text-query segmentation on
Figurines, Ramen, Teatime, and Waldo kitchen from
LERF~\citep{kerr2023lerf}, following the SFS evaluation
protocol~\citep{xiong2026sfs}.
For each query, we render the lifted features from the
evaluation camera and compare them with the query's
text embedding to obtain a segmentation mask.
Both representations use the same query evaluator,
reference masks, and threshold rules.
Table~\ref{tab:lerf-main} reports mean IoU and mean accuracy
across the four scenes.
\\OV-Foam improves both metrics for all four methods,
extending the advantage from 3D point labels
to rendered segmentation masks. Together, these results show that PowerFoam's advantage
extends beyond correct upstream labels and a single
lifting method.\enlargethispage{2\baselineskip} The earlier measurements identify differences in how
the representations distribute and combine image evidence.
The CLIP feature benchmarks establish that replacing
Gaussians with PowerFoam can turn those representation
differences into better open-vocabulary segmentation, even when the
input features remain imperfect.

\section{Limitations and conclusion}
Our ScanNet experiments cover ten scenes and initialize both representations from the same point cloud through the SfM initialization pipeline. For this benchmark, the supplied point cloud is the ground-truth geometry provided by ScanNet rather than an image-derived SfM reconstruction. The ground-truth lifting experiment isolates errors that remain with correct image labels, but does not establish an upper bound over all lifting methods. Our fitting-gap analysis concerns feature reconstruction loss rather than segmentation accuracy, and high purity can still reflect consistently incorrect observations. The observed advantages therefore depend on the reconstruction, input features, and evaluation setting.

We investigated how the choice of reconstruction primitives affects feature lifting. Under matched reconstruction budgets, PowerFoam has fewer contributors per ray and lower measured dispersion than 3DGS, while placing more evaluated points on primitives with high-purity evidence. Its one-shot estimate is closer to its least-squares reference, and lifting correct image labels produces more accurate 3D segmentation. These advantages extend to real image features, adapting existing lifting methods to PowerFoam improves the tested ScanNet and LERF segmentation results.
Together, these findings establish representation choice as a practical way to improve semantic lifting without requiring more accurate upstream image features.

\label{page:mainend}
\clearpage
\subsection*{AI use statement}

In this work, we used generative AI tools for drafting and editing sections of the paper.
We also used AI tools to find related works beyond what our manual research found. We used AI models to review and critique our paper and used its feedback to improve our work. AI coding agents were used for parts of the implementation and debugging. All generated text, code and suggested papers were reviewed thoroughly by the authors. In all cases we used AI tools as \emph{tools} to assist our work, never to replace or automate full aspects of the research process.

\subsection*{Reproducibility statement}

With our paper, we release the full source code required to reproduce our results.
We will upload the reconstructed 3DGS and Powerfoam scenes and the lifted feature sets used as input for
our segmentation. We will supply a simple Python script that reproduces the segmentations that lead to the results.

\bibliography{references}
\bibliographystyle{plainnat}
\clearpage
\appendix
\section{Dispersion and Contribution count}
\label{app:matched}

\begin{table}[h]
\centering
\small
\setlength{\tabcolsep}{4pt}
\caption{\textbf{Per-scene matched-budget results on ScanNet.}
Both representations use the same primitive count $P$ within each scene.
PSNR uses all views. Contributor count $k_i$ and dispersion $\beta_i$
use twelve views per scene. The final row averages the scene summaries,
including the scene medians of $\beta_i$.
Bold indicates higher PSNR or lower contributor count and dispersion.}
\label{tab:matched-full}
\begin{tabular}{lrrrrrrr}
\toprule
& & \multicolumn{2}{c}{PSNR (dB)}
  & \multicolumn{2}{c}{Mean $k_i$}
  & \multicolumn{2}{c}{Median $\beta_i$} \\
\cmidrule(lr){3-4}
\cmidrule(lr){5-6}
\cmidrule(lr){7-8}
Scene & $P$ & Foam & 3DGS & Foam & 3DGS & Foam & 3DGS \\
\midrule
scene0000\_00 & 81{,}369  & \textbf{21.84} & 19.74 & \textbf{2.32} & 31.52 & \textbf{0.0053} & 1.8443 \\
scene0062\_00 & 51{,}610  & \textbf{27.31} & 23.16 & \textbf{2.02} & 33.90 & \textbf{0.0000} & 1.0508 \\
scene0070\_00 & 109{,}380 & 28.13 & \textbf{29.46} & \textbf{2.69} & 34.11 & \textbf{0.0318} & 2.1095 \\
scene0097\_00 & 72{,}007  & 31.94 & \textbf{34.44} & \textbf{1.91} & 33.26 & \textbf{0.0000} & 1.1392 \\
scene0140\_00 & 372{,}941 & \textbf{27.25} & 26.62 & \textbf{2.76} & 34.24 & \textbf{0.0092} & 1.6057 \\
scene0200\_00 & 83{,}291  & \textbf{26.06} & 21.09 & \textbf{2.00} & 34.82 & \textbf{0.0000} & 1.1613 \\
scene0347\_00 & 67{,}984  & \textbf{25.24} & 20.56 & \textbf{2.13} & 31.35 & \textbf{0.0001} & 1.2667 \\
scene0400\_00 & 155{,}959 & \textbf{26.60} & 21.48 & \textbf{1.96} & 36.26 & \textbf{0.0000} & 1.5230 \\
scene0590\_00 & 222{,}957 & 31.67 & \textbf{32.50} & \textbf{2.73} & 33.52 & \textbf{0.0035} & 1.7131 \\
scene0645\_00 & 352{,}477 & \textbf{24.04} & 23.15 & \textbf{2.66} & 33.36 & \textbf{0.0302} & 2.1408 \\
\midrule
Mean & 156{,}998 & \textbf{27.01} & 25.22
     & \textbf{2.32} & 33.63 & \textbf{0.0080} & 1.5554 \\
\bottomrule
\end{tabular}
\end{table}

\section{ScanNet evaluation protocol}
\label{app:scannet-eval}

We follow OpenGaussian's ten-scene point-level evaluation
setup~\citep{wu2024opengaussian}. For our matched reconstructions,
both representations start from the same ground-truth vertices,
with one primitive per vertex. Primitive centers and counts
remain fixed during reconstruction, while the remaining
reconstruction parameters are optimized. We then hold the
reconstruction fixed during feature lifting.

We assign semantic labels by maximum cosine similarity to the
class text embeddings and report mIoU and mAcc for the 19-,
15-, and 10-class evaluations. Within each scene, these metrics
average class IoU and class recall over the valid ground-truth
classes present. Final scores average the scene results.

\subsection{Dipole-aware point assignment}
\label{app:dipole-assignment}

For OV-Foam, point ownership uses the reconstructed dipole geometry.
A fixed initialization center need not lie on the occupied side
of its primitive's displaced surface. For each evaluation point $q$,
we take the 64 Euclidean-nearest supported primitive centers and rank
them by power distance $\|q-p_j\|^2-a_j^2$. We select the first
candidate satisfying $(q-p_j)\cdot n_j\leq h_j(q)$, where $h_j(q)$
is the interpolated dipole height at the point's tangent-plane
coordinates. This test checks the occupied side of the dipole surface.

If no candidate passes, we use the minimum power distance in the same
pool. The fallback preserves assignment coverage, so the dipole test
does not discard points. Power distances are evaluated in double
precision. The finite candidate search approximates global
power-distance ranking.

\subsection{Evaluation without opacity culling}
\label{app:opacity-culling}

OpenGaussian excludes GT points associated with Gaussians whose
opacity is below $0.1$. The Del.\% column in
Table~\ref{tab:scannet} reports the fraction of labeled points
excluded before scoring. Rates marked $^{\ddagger}$ come from
our matched reproductions. This exclusion changes which points contribute to the metrics.
To compare both representations on the same evaluation support,
Table~\ref{tab:scannet-unfiltered} retains all valid labeled
ground-truth points, keeping the reconstructions and lifted
features fixed. Removing opacity culling widens OV-Foam's advantage
over the matched Gaussian baselines. OV-Foam also achieves
higher mIoU and mAcc for every method across the class set
evaluations. These results show that its advantage persists and
increases when low-opacity points are included in scoring.

\begin{table}[t]
\centering\small
\setlength{\tabcolsep}{3.4pt}
\caption{\textbf{ScanNet results under opacity-filtered evaluation.}
Published Gaussian scores are paired with our OV-Foam adaptations.
Bold marks the better score within each pair.
Green parentheses show OV-Foam minus 3DGS in percentage points.
Del.\% denotes excluded labeled points.
$^{\ddagger}$Rates are measured on our matched reproductions,
not reported by the cited papers.
Table~\ref{tab:scannet-unfiltered} reports results without filtering.}
\label{tab:scannet}\label{tab:scannet-paired}\label{tab:scannet-main}
\begingroup
\newcommand{\foamdelta}[2]{%
  \textcolor{green!50!black}{($#2$)}\,#1}
\resizebox{\textwidth}{!}{%
\begin{tabular}{llrrrrrrr}
\toprule
& & \multicolumn{2}{c}{19 classes}
  & \multicolumn{2}{c}{15 classes}
  & \multicolumn{2}{c}{10 classes} & \\
\cmidrule(lr){3-4}\cmidrule(lr){5-6}\cmidrule(lr){7-8}
Method & Representation
& mIoU & mAcc & mIoU & mAcc & mIoU & mAcc & Del.\%\\
\midrule
VALA~\citep{wang2025vala}
& 3DGS
& 32.11 & 50.05 & 35.10 & 54.77
& \textbf{46.21} & \textbf{65.61} & 14.16$^{\ddagger}$\\
& OV-Foam
& \foamdelta{\textbf{33.68}}{+1.57}
& \foamdelta{\textbf{51.18}}{+1.13}
& \foamdelta{\textbf{37.15}}{+2.05}
& \foamdelta{\textbf{56.08}}{+1.31}
& \foamdelta{45.68}{-0.53}
& \foamdelta{65.39}{-0.22}
& \textbf{0.00}\\
\addlinespace[2pt]
Occam's LGS~\citep{cheng2024occam}
& 3DGS
& 31.93 & 48.93 & 34.25 & 53.71
& 45.16 & 64.39 & 19.28$^{\ddagger}$\\
& OV-Foam
& \foamdelta{\textbf{34.19}}{+2.26}
& \foamdelta{\textbf{52.35}}{+3.42}
& \foamdelta{\textbf{37.59}}{+3.34}
& \foamdelta{\textbf{57.47}}{+3.76}
& \foamdelta{\textbf{46.47}}{+1.31}
& \foamdelta{\textbf{66.65}}{+2.26}
& \textbf{0.10}\\
\addlinespace[2pt]
SFS~\citep{xiong2026sfs}
& 3DGS
& 33.33 & 51.35 & 36.43 & 55.38
& 44.74 & 63.53 & 12.66$^{\ddagger}$\\
& OV-Foam
& \foamdelta{\textbf{34.01}}{+0.68}
& \foamdelta{\textbf{51.76}}{+0.41}
& \foamdelta{\textbf{37.61}}{+1.18}
& \foamdelta{\textbf{57.45}}{+2.07}
& \foamdelta{\textbf{46.92}}{+2.18}
& \foamdelta{\textbf{66.66}}{+3.13}
& \textbf{0.10}\\
\addlinespace[2pt]
LUDVIG~\citep{marrie2025ludvig}
& 3DGS
& 33.90 & 51.40 & 37.40 & 57.20
& 46.40 & 66.20 & 15.92$^{\ddagger}$\\
& OV-Foam
& \foamdelta{\textbf{34.59}}{+0.69}
& \foamdelta{\textbf{52.15}}{+0.75}
& \foamdelta{\textbf{38.12}}{+0.72}
& \foamdelta{\textbf{57.86}}{+0.66}
& \foamdelta{\textbf{47.60}}{+1.20}
& \foamdelta{\textbf{67.05}}{+0.85}
& \textbf{0.11}\\
\bottomrule
\end{tabular}%
}
\endgroup
\end{table}

\section{Full Oracle table}
\label{app:full-oracle}
\begin{table}[h]\centering\setlength{\tabcolsep}{4pt}
\caption{\textbf{Per-scene lifting of ground-truth image semantics on ScanNet.}
Point accuracy and mIoU (\%) when each valid pixel carries its ground-truth class text embedding,
using ScanNet's released 2D label masks. Geom UB is the geometric upper bound: each primitive is
assigned the majority ground-truth class of its evaluated points.
Bold marks the higher score within each pair.
$C$ is the number of ground-truth classes present, and $N$ is the number of scored points. This shows 3DGS doesn't lose due to lack of capacity but due to the solve errors.
The Mean row is the equal-weight average over scenes and matches Table~\ref{tab:oracle}.}
\label{tab:gt-lifting-perscene}
\vspace{1mm}
\begingroup
\newcommand{\dgain}[1]{\textcolor{green!50!black}{(#1)}}
\newcommand{\dloss}[1]{\textcolor{red!60!black}{(#1)}}
\resizebox{\textwidth}{!}{
\begin{tabular}{lrrrrrrrrrrr}
\toprule
& & & \multicolumn{3}{c}{Point accuracy}
    & \multicolumn{3}{c}{mIoU}
    & \multicolumn{3}{c}{Geom UB (mIoU)} \\
\cmidrule(lr){4-6}\cmidrule(lr){7-9}\cmidrule(lr){10-12}
Scene & $C$ & $N$
& 3DGS & OV-Foam & $\Delta$
& 3DGS & OV-Foam & $\Delta$
& 3DGS & OV-Foam & $\Delta$ \\
\midrule
scene0000\_00 & 14 & 54{,}480  & 92.98 & \textbf{96.10} & \dgain{+3.12} & 83.61 & \textbf{91.27} & \dgain{+7.66} & 97.00 & \textbf{97.81} & \dgain{+0.81} \\
scene0062\_00 & 6  & 41{,}873  & 95.85 & \textbf{96.50} & \dgain{+0.65} & 86.53 & \textbf{87.82} & \dgain{+1.29} & \textbf{98.94} & 97.50 & \dloss{-1.44} \\
scene0070\_00 & 9  & 69{,}922  & 94.10 & \textbf{96.77} & \dgain{+2.67} & 83.30 & \textbf{89.89} & \dgain{+6.59} & \textbf{98.69} & 97.92 & \dloss{-0.77} \\
scene0097\_00 & 7  & 52{,}656  & 95.41 & \textbf{97.34} & \dgain{+1.93} & 88.37 & \textbf{93.06} & \dgain{+4.69} & \textbf{99.33} & 97.81 & \dloss{-1.52} \\
scene0140\_00 & 8  & 279{,}359 & 91.82 & \textbf{94.11} & \dgain{+2.29} & 86.03 & \textbf{90.80} & \dgain{+4.77} & \textbf{99.19} & 98.95 & \dloss{-0.24} \\
scene0200\_00 & 8  & 65{,}082  & 93.65 & \textbf{96.07} & \dgain{+2.42} & 83.11 & \textbf{89.59} & \dgain{+6.48} & \textbf{99.20} & 97.69 & \dloss{-1.51} \\
scene0347\_00 & 7  & 48{,}063  & 95.06 & \textbf{97.07} & \dgain{+2.01} & 88.44 & \textbf{93.10} & \dgain{+4.66} & \textbf{99.34} & 99.05 & \dloss{-0.29} \\
scene0400\_00 & 7  & 93{,}930  & 94.94 & \textbf{97.26} & \dgain{+2.32} & 85.36 & \textbf{91.23} & \dgain{+5.87} & \textbf{99.57} & 98.87 & \dloss{-0.70} \\
scene0590\_00 & 14 & 153{,}151 & 93.91 & \textbf{95.78} & \dgain{+1.87} & 84.74 & \textbf{88.85} & \dgain{+4.11} & \textbf{98.99} & 97.28 & \dloss{-1.71} \\
scene0645\_00 & 16 & 253{,}034 & 94.76 & \textbf{96.71} & \dgain{+1.95} & 84.55 & \textbf{90.24} & \dgain{+5.69} & \textbf{98.96} & 98.46 & \dloss{-0.50} \\
\midrule
Mean & -- & -- & 94.25 & \textbf{96.37} & \dgain{+2.12} & 85.40 & \textbf{90.59} & \dgain{+5.19} & \textbf{98.92} & 98.13 & \dloss{-0.79} \\
\bottomrule
\end{tabular}}
\endgroup
\end{table}

\section{Proof of the fitting-gap identity}
\label{app:proofs}
Let $A$ contain the rendering weights, $B$ the observed
image features, and $X$ the primitive features.
The rendering loss is
\begin{equation}
\mathcal{L}(X)
= \frac12 \|AX-B\|_F^2
= \frac12 \sum_{i=1}^{R}\sum_{f=1}^{d}
  \bigl((AX)_{if}-B_{if}\bigr)^2,
\end{equation}
where $R$ is the number of rays and $d$ is the feature
dimension. The subscript $F$ denotes the Frobenius norm,
which sums squared entries of a matrix as shown above.

Let $\widehat X$ be an exact, unconstrained least-squares
minimizer and $X'$ the one-shot estimate, both using the
same $A$ and $B$. Then
\begin{equation}
\boxed{
\mathcal{L}(X')-\mathcal{L}(\widehat X)
= \frac12 \|A(X'-\widehat X)\|_F^2.
}
\label{eq:appendix-fitting-gap}
\end{equation}

To prove this, write the one-shot rendering residual as
\[
AX'-B = (A\widehat X-B)+A(X'-\widehat X).
\]
Expanding its squared norm gives
\begin{align}
\mathcal{L}(X')-\mathcal{L}(\widehat X)
&= \frac12 \|A(X'-\widehat X)\|_F^2 \\
&\quad+
\sum_{i=1}^{R}\sum_{f=1}^{d}
(A\widehat X-B)_{if}
\bigl[A(X'-\widehat X)\bigr]_{if}.
\end{align}
The final term is zero because the least-squares
normal equations give
$A^\top(A\widehat X-B)=0$.
Thus the fitting gap equals half the squared difference
between the image features rendered by the two estimates.

The identity allows incomplete absorption and does not
require a unique least-squares solution.
It measures agreement between rendered features,
not correctness of the resulting semantic labels.
For a numerically computed reference, the equality
holds up to its remaining optimization error.

\section{Qualitative figures}
\label{app:qualitative}
\begin{figure}[h]
\centering\includegraphics[width=.93\linewidth,height=.78\textheight,keepaspectratio]{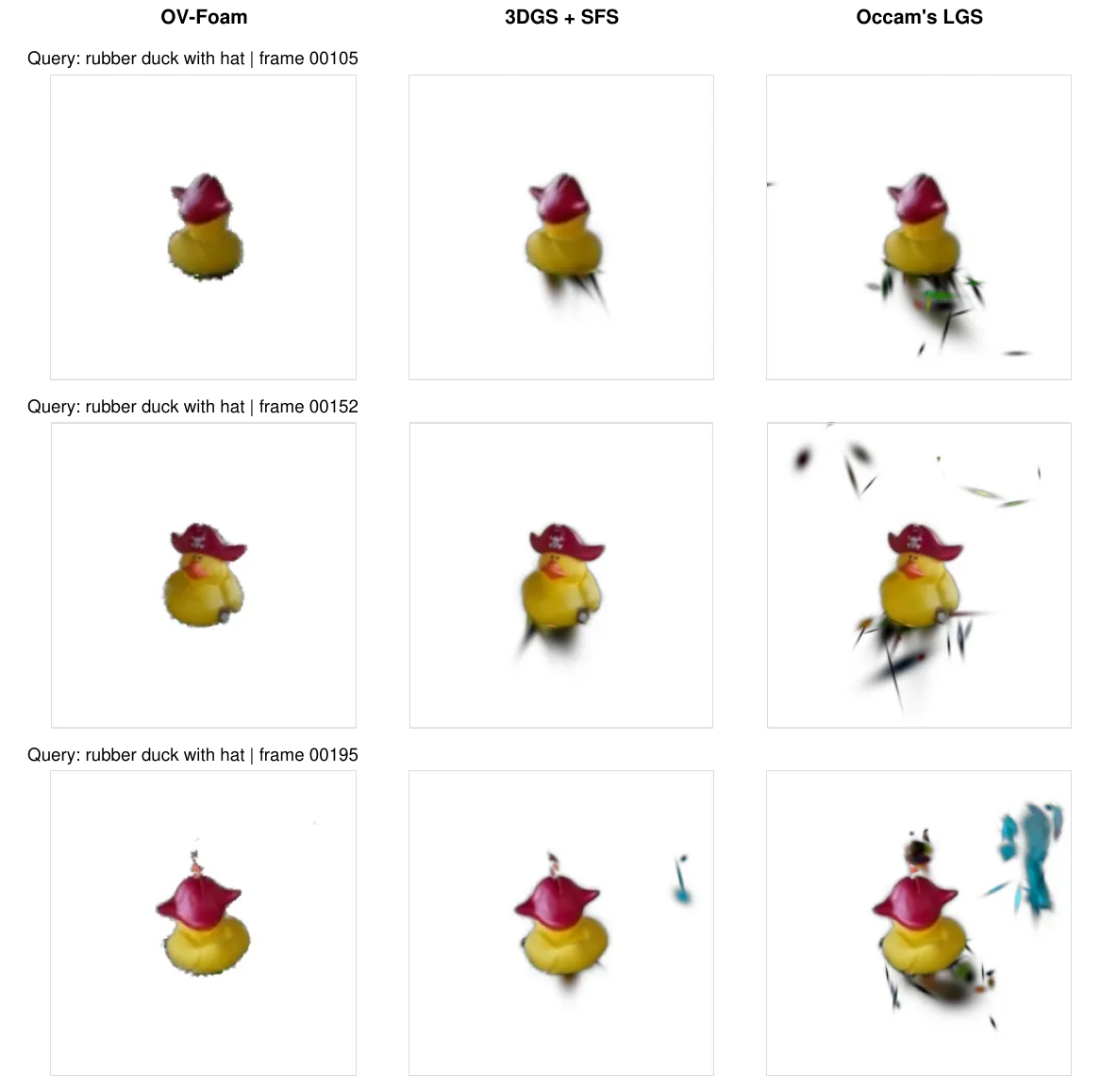}
\caption{Object extraction for ``rubber duck with hat, frames 00105, 00152 and 00195. OV-Foam uses SFS here}
\end{figure}

\begin{figure}[h]
\centering\includegraphics[width=.98\linewidth,height=.83\textheight,keepaspectratio]{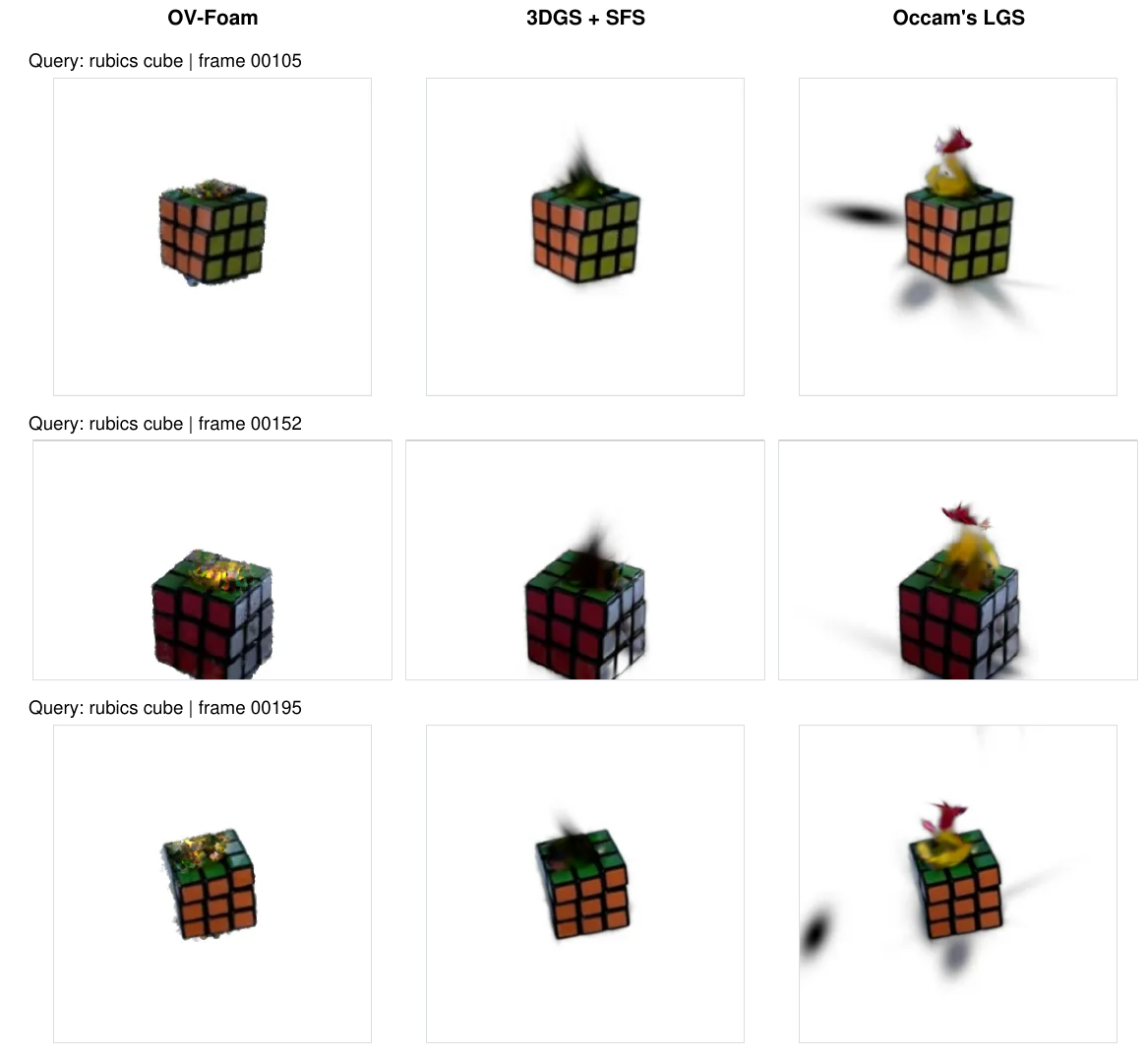}
\caption{Object extraction for ``rubics cube, frames 00105, 00152 and 00195. OV-Foam uses SFS here}
\end{figure}

\begin{figure}[h]
\centering\includegraphics[width=.98\linewidth,height=.83\textheight,keepaspectratio]{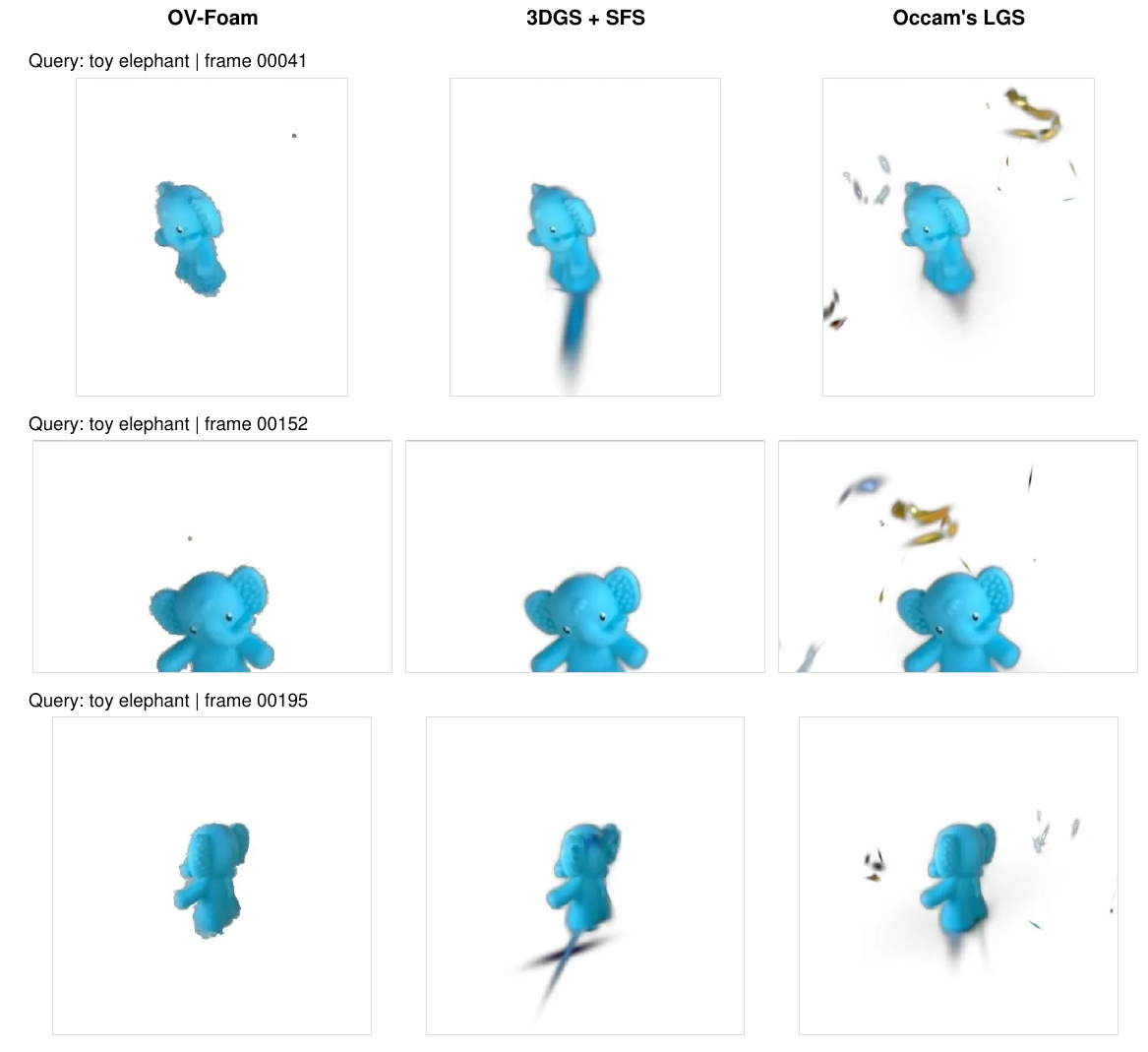}
\caption{Object extraction for ``toy elephant, frames 00041, 00152 and 00195. OV-Foam uses SFS here}
\end{figure}

\begin{figure}[h]
\centering\includegraphics[width=.98\linewidth,height=.83\textheight,keepaspectratio]{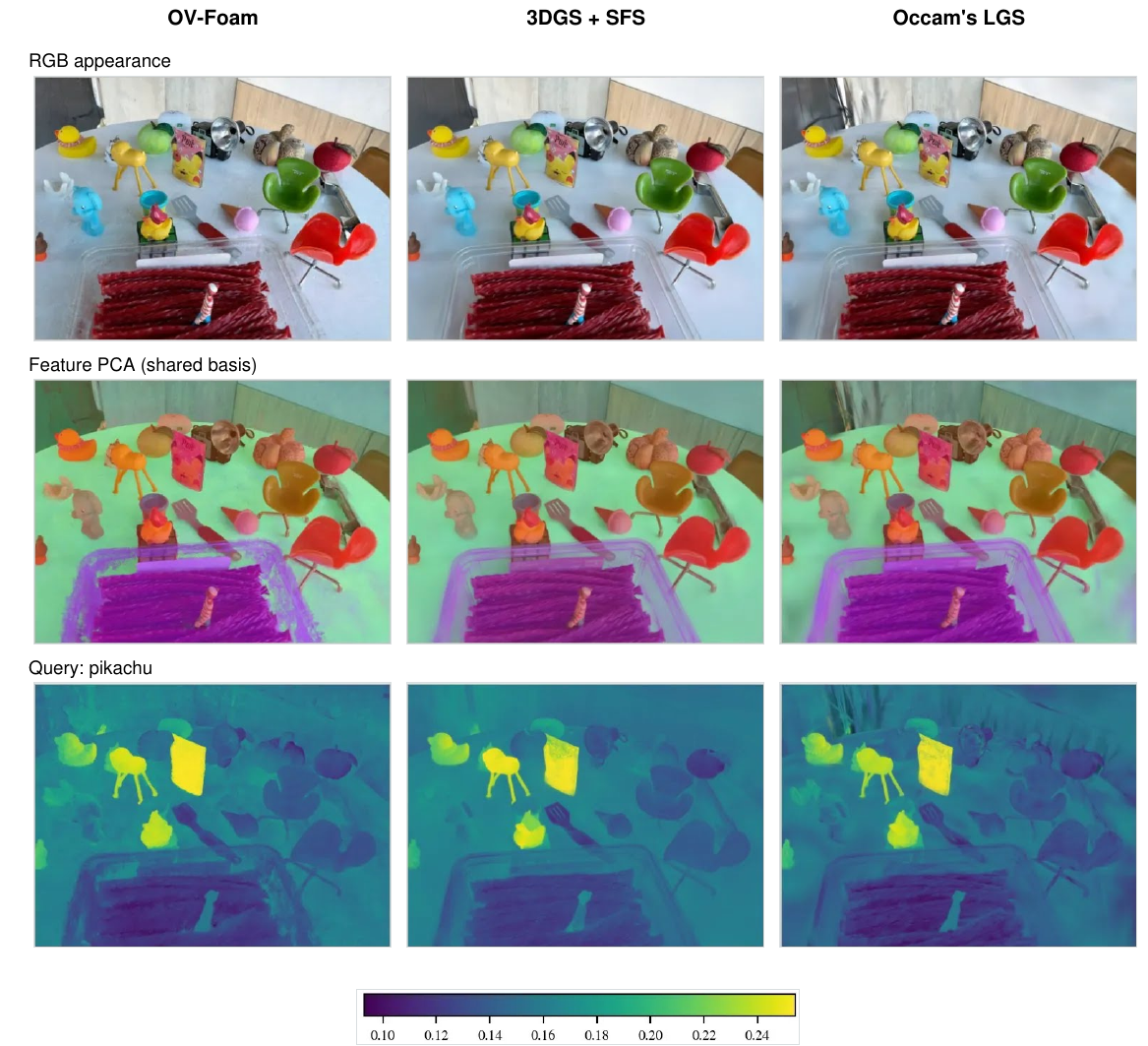}
\caption{RGB appearance, shared-basis feature PCA and cosine similarity to ``pikachu, frame 00041. OV-Foam uses SFS here}
\end{figure}

\begin{figure}[h]
\centering\includegraphics[width=.98\linewidth,height=.83\textheight,keepaspectratio]{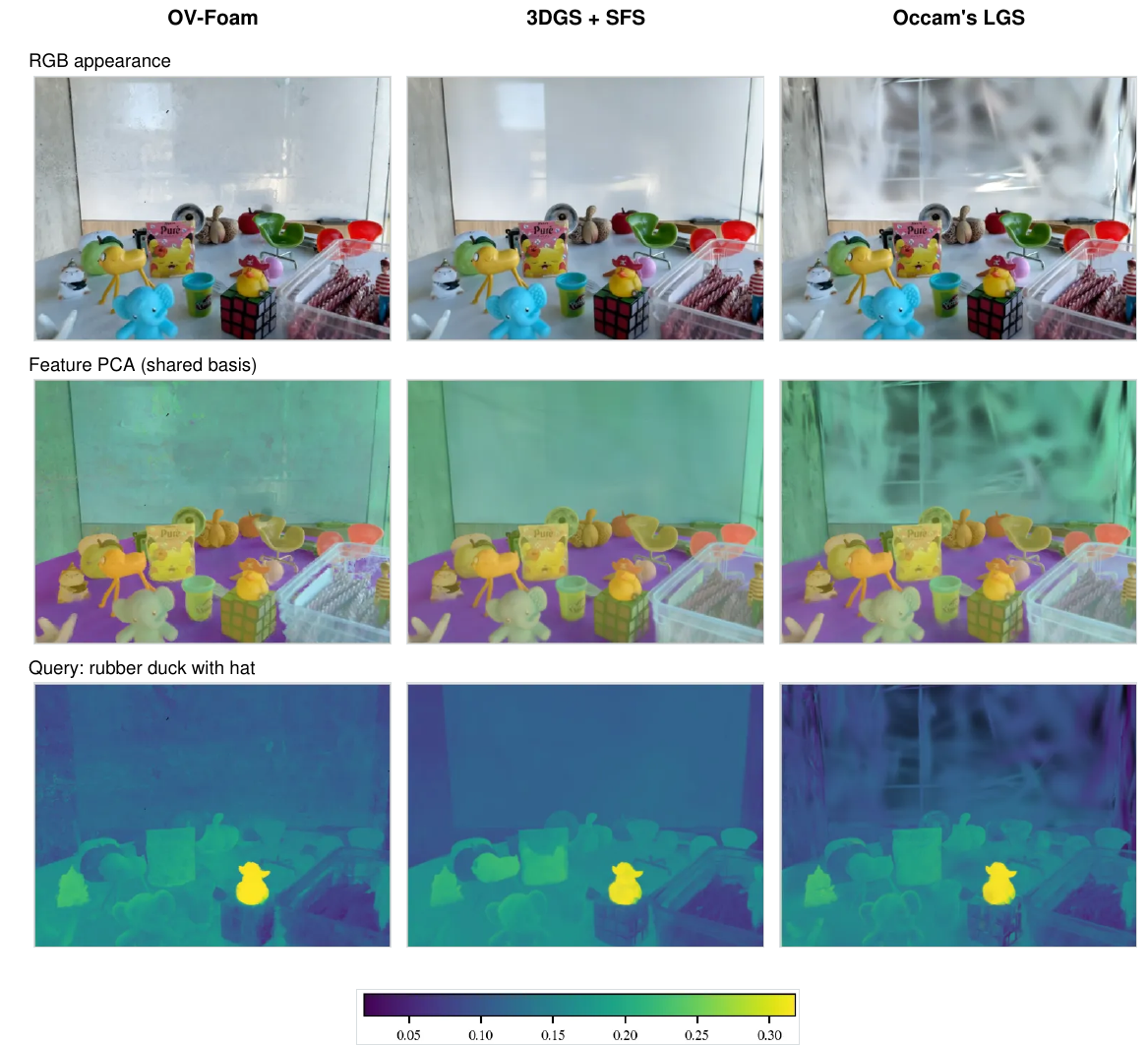}
\caption{RGB appearance, shared-basis feature PCA and cosine similarity to ``rubber duck with hat, frame 00152. The supplied overlay uses alpha 0.55. OV-Foam uses SFS here}
\end{figure}

\begin{figure}[h]
\centering\includegraphics[width=.98\linewidth,height=.83\textheight,keepaspectratio]{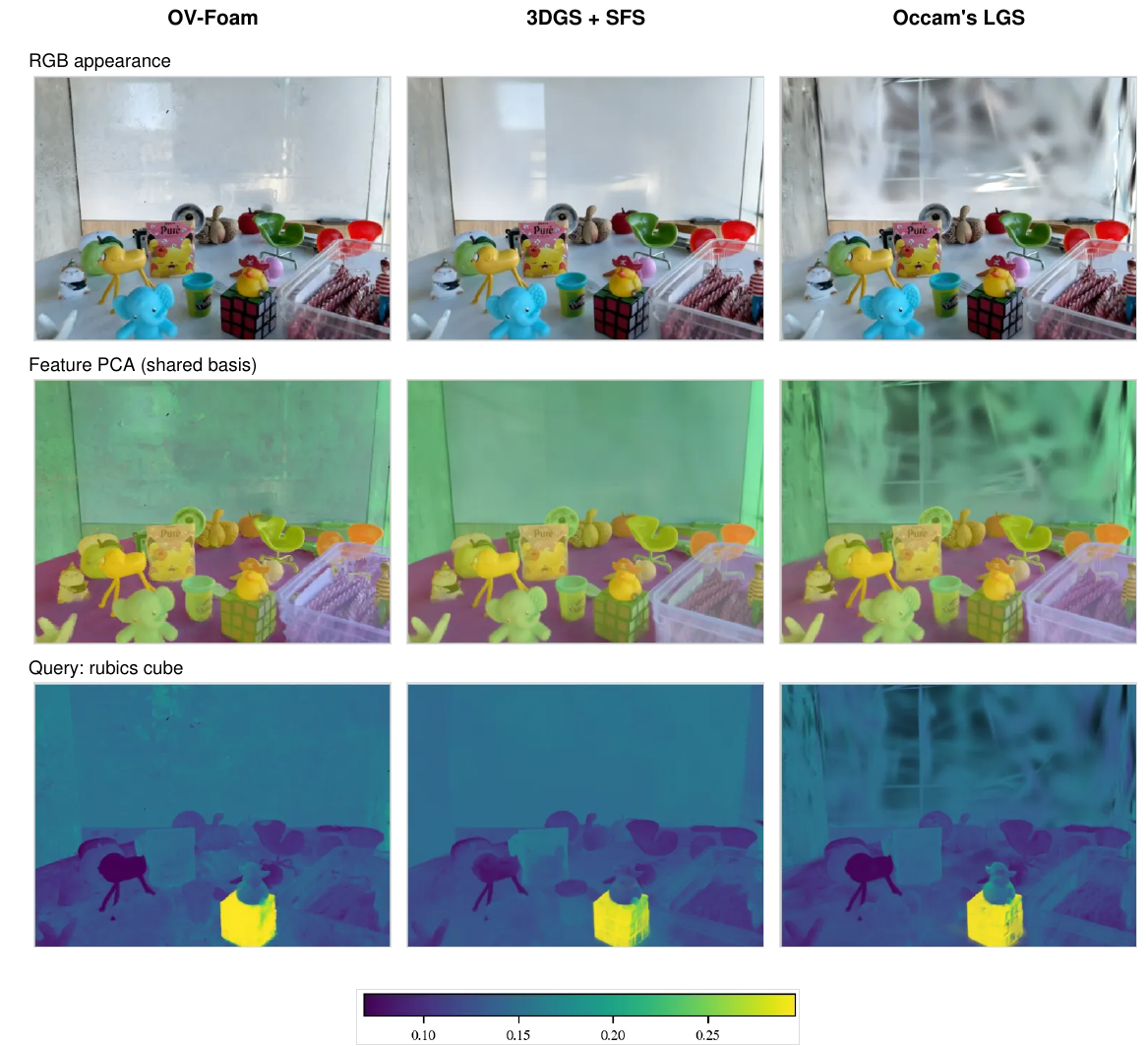}
\caption{RGB appearance, shared-basis feature PCA and cosine similarity to ``rubics cube, frame 00152. The supplied overlay uses alpha 0.55. OV-Foam uses SFS here}
\end{figure}
\clearpage

\end{document}